\documentclass[11pt]{article}

\usepackage[utf8]{inputenc}
\usepackage[T1]{fontenc}
\usepackage{amsmath,amssymb}
\usepackage{graphicx}
\graphicspath{{figures/}{figures/}}
\usepackage{booktabs}
\usepackage{array}
\usepackage{multirow}
\usepackage{url}
\usepackage{placeins}   %
\usepackage[hidelinks]{hyperref}
\usepackage[textwidth=6.5in,textheight=9in]{geometry}
\usepackage{natbib}
\usepackage{xcolor}
\usepackage{enumitem}

\ifdefined\anonymisedtrue\else
  \expandafter\newif\csname ifanonymised\endcsname \anonymisedfalse
\fi
\newcommand{\commit}[1]{\ifanonymised[commit anonymised for review]\else\texttt{#1}\fi}
\newcommand{\repopath}[1]{\ifanonymised[path anonymised for review]\else\texttt{#1}\fi}

\hypersetup{pdfauthor={},pdfcreator={},pdfsubject={},pdfkeywords={}}
\title{Externalise the Memory:\\
A Pre-Registered Study of Where an Embodied Agent's\\
Experience Should Live}

\author{%
  Cheng-Yin Tu\\
  Nikon Research Corporation of America\\
  \texttt{chengyin.tu@nikon.com}%
}
\date{}

\begin{document}
\maketitle

\begin{abstract}
Should an embodied agent's accumulated experience be baked into its
policy weights, or kept external and retrieved at inference? We test
this directly. A memory-augmented teacher (a VLM that retrieves lessons,
with their episode images, from its own past attempts) produces trajectories on a
vision-conditioned single-arm reach task; from the \emph{same} trajectories
we distil LoRA student policies and evaluate them under matched
protocols, including one that re-attaches the teacher's frozen
retrieval buffer to a fixed set of student weights --- a clean parametric-%
vs-contextual-memory contrast on identical weights. Because the
headline result is a null, we pre-registered the hypotheses, arms,
tests, and analysis rules before training and before any $p$-value.
Distilling the memory-augmented behaviour into weights beat an
action-only control by only $+1$ pp (n.s.); the registered test proves
underpowered for its own minimum effect, so we withdraw --- rather than
refute --- the memory-in-weights claim. By contrast, attaching retrieval
to a \emph{fixed} set of student weights added $+34$ pp ($z=5.15$),
reaching the pooled memory-augmented teacher reference (49.3\%); the
pre-registered protocol contrast agrees ($+23$ pp, $z=3.36$); under a
pre-registered auto-reset sensitivity count the two are $+36$ and $+29$ pp.
An exploratory behavioural probe with pre-registered per-arm predictions is
consistent with a mechanism: behavioural distillation transfers the
\emph{marginal} correction (a static prior every arm adopts) but not the
\emph{conditional}, situation-specific correction that prompt-retrieved
memory supplies. The recipe that follows is to externalise the memory: in
follow-up controls registered before they ran, the same frozen buffer
lifts the base model with no adapter to 48\% (sensitivity 48\%) and each
of the other three adapters from 19--26\% to 53--55\% (sensitivity 17--21\%
to 50--55\%), so the retrieval effect does not
depend on the adapter, and the adapter adds nothing detectable once the
buffer is attached. On a second, harder task in the same family, the
retrieval advantage points the same direction but falls short of
significance ($+6$ pp, 95\% CI $[-4.5, +16.4]$); exploratory diagnostics tie
the attenuation to memory coverage and a distance-gated conversion
threshold, suggesting the effect scales with the amount of retrievable
success experience.
\end{abstract}

\section*{1. Introduction}

\subsection*{Opening problem}

When an embodied agent accumulates experience, where should that
experience live: folded into the policy's parameters, or kept outside
and retrieved when needed? The question is old and unresolved. The
parametric view is seductive: a fluent agent, like a fluent human,
should not have to look things up mid-task; skill that has been
practised enough becomes automatic, and automaticity feels like
something written into the substrate. The contextual view is
pragmatic: retrieval systems that keep experience external and look it
up have repeatedly matched or beaten parametric memorisation in
language tasks. For embodied agents the question is structural: a
retrieval context grows with every experience the agent accumulates,
while its weights do not, so ``bake it into the weights'' is
architecturally convenient as well as philosophically appealing. Whether
it \emph{works} is an empirical question that, to our knowledge, has not
been asked for embodied skill under controlled conditions.

\subsection*{Why this test bed answers it}

We ask it with a design that isolates the variable of interest. A
memory-augmented teacher (a VLM that retrieves lessons, with their
episode images, from its own past attempts) produces trajectories on a vision-conditioned
single-arm reach task (an Isaac Lab bimanual rig with the right arm
frozen). From those \emph{same} trajectories we distil student policies
whose training targets differ only in whether retrieved-lesson content
is present, and we evaluate them under matched protocols, including
one that re-attaches inference-time retrieval to a fixed set of student
weights. The teacher's experience buffer is thus routed to the student
either as distillation targets (lesson gist in text) or as retrieval
context at inference (lesson text with its episode image). The critical pair shares identical adapter weights and differs
only in protocol: memory supplied by baking it into those weights
during distillation, versus memory supplied through a retrieval context
at inference. This makes ``parametric vs contextual memory'' a
controlled contrast on one axis.

\subsection*{Pre-registration as a stance}

Because the headline outcome is a null, and nulls are the most easily
disbelieved result in machine learning, we pre-registered in two
locked layers. The hypotheses, comparison arms, primary tests, and
their directionality were registered \emph{before any adapter trained}; the
final analysis rules (the pairing-integrity gate and its mechanical
test-selection fallback) were locked \emph{after collection but before any
$p$-value was computed}. That locked rule kept the raw success counts,
which were visible in the collection logs, out of the test-selection
decision. The pre-registration lets a reader take the null as evidence
rather than as a failure to find an effect we wanted. Every result below
is labelled confirmatory or exploratory against that registered plan.

\subsection*{Contributions}

\begin{enumerate}
\item \textbf{A design criterion for where experience should live}: marginal
knowledge (the correction that holds on average across situations)
transfers into weights, while conditional knowledge (the correction
that depends on the current situation) belongs in context, because
weights can only approximate a conditional function with a constant
whereas context conditions on it for free. We name and operationalise
this marginal/conditional distinction; the split we observe is
consistent with it. \emph{(This is the transferable takeaway; the results
below are consistent with it.)}
\item \textbf{A pre-registered null on parametric memory} (T1): distilling the
memory-augmented teacher's behaviour into student weights beat an
action-only control by $+1$ pp (n.s.; auto-reset sensitivity $-2$ pp), and we withdrew the
memory-in-weights claim per the registered falsification clause. The
withdrawal is scoped to this recipe (single-round SFT + composable-KTO), this
task, this data scale, and this adapter capacity (LoRA rank-16). A
no-training diagnostic (\S4.4b) finds no detectable conditional trace at
round 1; whether more data of the same kind would grow one remains open.
\item \textbf{A reversal}: on a \emph{fixed} set of student weights, attaching
inference-time retrieval added $+34$ pp (identical-weights contrast,
C\_retrieval vs A\_ctrl\_rat, $z=5.15$, exploratory); the pre-registered
protocol contrast (T4, C\_retrieval vs B\_main, each with its own
adapter) agrees at $+23$ pp ($z=3.36$, confirmatory); sensitivity $+36$ /
$+29$ pp. Retrieval reaches the pooled memory-teacher reference (49.3\%).
\item \textbf{A mechanism reading} (exploratory R1 $\Delta$X probe with
pre-registered per-arm predictions, Amendment 14 \S14.5): the data are
consistent with distillation transferring the \emph{marginal}
distribution shift (a static prior all arms adopt regardless of
situation) but not the \emph{conditional} structure (the
situation-specific correction that retrieval supplies). This selects the
registered H\_behavior branch as an exploratory reading: the
correction's carrier is the action distribution, not the memory content.
\end{enumerate}

\subsection*{Summary}

The two headline results are one finding: distillation moves the
average, and retrieval supplies the situation. The recipe is therefore to
externalise the memory; with the buffer attached, the adapter adds nothing
detectable (\S4.2b).
\section*{2. Related Work}

\paragraph{Parametric vs.\ retrieved knowledge in language models.} A large line
compares injecting knowledge into weights (fine-tuning) against keeping
it external and retrieving it (RAG~\citep{rag2020}), repeatedly finding
retrieval competitive or superior, though not uniformly: for LM-based
world modelling fine-tuning wins most settings while retrieval is more
data-efficient~\citep{worldmodelft2026}, and a holistic comparison of memory
substrates finds none dominant~\citep{harnessmem2026}; recent work
on self-editing / self-adapting weights (e.g.\ SEAL~\citep{seal2025}) pushes the
parametric pole. Our contribution is the embodied, controlled analogue
of this comparison: same weights, memory supplied two ways.

\paragraph{Agent memory and experiential learning.} Voyager~\citep{voyager2023},
ExpeL~\citep{expel2024}, and
Agent-Workflow-Memory~\citep{awm2024} accumulate and reuse experience as external,
retrievable artifacts (skills, lessons, workflows); this is the
external-memory pole our T4 result supports. A parallel line folds agent
experience into weights instead: ASCENT~\citep{ascent2026} self-distils
verified trajectories into the weights at test time, EDGE~\citep{edge2026}
treats retrieved experience as a training-time scaffold and distils its
gains into the policy so that retrieval is not needed at inference, and
UniMem~\citep{unimem2026} routes recurring patterns to weights and sparse
ones to an episodic store. Closest to our question,
\citet{routing2026} compare fine-tuning and prompt retrieval on the
\emph{same} experience items for GUI agents and find that the destination
depends on the item: state facts and procedures do better in context,
locators and lessons in the weights, predicted by the item's recurrence and
state-conditionality. Their state-conditionality rule parallels our
marginal/conditional split; their lessons-in-weights result differs from
ours, where retrieved lessons (with their episode images) helped only in context. We differ in
isolating, under pre-registration and in an embodied teacher-to-student
setting with identical weights in both arms, whether that experience can
instead be consolidated into weights.

\paragraph{Memory for VLAs and embodied policies.} A fast-moving line
gives embodied policies explicit memory. Some keep it \emph{external and
retrieved}: ExpReS-VLA~\citep{expresvla2025} specialises VLAs through
experience replay and retrieval, Retrieve-then-Steer~\citep{retrievethensteer2026}
keeps an online success memory for test-time adaptation, and
STRAP~\citep{memmel2024strap} / RAEA~\citep{zhu2024raea} retrieve
sub-trajectories or augment embodied agents; MemoryVLA~\citep{memoryvla2025}
builds a perceptual-cognitive memory into the architecture, and
RoboMME~\citep{robomme2026} benchmarks memory for generalist policies;
for in-context retrieval in VLAs, no retrieval method dominates across
settings~\citep{whenretrieval2026}, and test-time retrieval helps most where
the frozen policy lacks the relevant action prior~\citep{reusebefore2026}.
Others fold experience \emph{into weights}: RECAP / $\pi^{*}_{0.6}$~\citep{pistar06_2025}
learns from experience via advantage-conditioned post-training. Our
contribution is orthogonal: rather than proposing a memory system, we hold
the \emph{same} retrieved memory fixed and ask which substrate carries it,
under pre-registration.

\paragraph{VLA post-training and RL.} Recent vision-language-action work
refines pretrained policies with RL and preference signals (e.g.\
SimpleVLA-RL~\citep{simplevlarl2025} and related); our students are distilled and KTO-refined VLAs,
and our null speaks to what such post-training does and does not move
into parameters.

\paragraph{Lifelong / continual LoRA.} CORAL~\citep{luo2026coral}, TAIL~\citep{tail2023}, and related study adapter-%
based continual adaptation; our marginal/conditional distinction
offers a lens on \emph{what} such adapters can absorb: a static prior
readily, but not a conditional map (at this recipe and capacity).

\paragraph{Reasoning distillation.} Distilling Step-by-Step~\citep{distillstepbystep2023}, Embodied-CoT~\citep{embodiedcot2024}, and
fast/slow thinking-action work distil chain-of-thought into cheaper
students; our rationale-in-target arms (A\_ctrl\_rat, B\_main) are
instances of this, and our finding that self-produced rationale at
inference did not help ($+1$ pp) while retrieved content did ($+23$ pp)
qualifies when rationale externalisation pays.

\paragraph{Pre-registration in ML.} We adopt pre-registration~\citep{nosek2018preregistration,bertinetto2021prereg} as a methodological stance: registering hypotheses,
tests, and analysis rules before seeing outcomes is what converts a
null from a non-result into evidence. We do not claim primacy; we claim
only that the study is explicitly pre-registered with a full,
amendment-tracked chain, which is uncommon in embodied learning and is
what licenses the null's interpretation.

Every design choice below is traceable to a pre-registration section or a
numbered amendment; the full amendment chain is in Appendix~A1.

\section*{3. Methods}

\subsection*{3.1 Task and environment}

We use a vision-conditioned single-arm reach task (L0a-Left) in an
Isaac Lab~\citep{orbit2023} bimanual manipulation rig with the right arm held frozen
throughout, a controlled setting in which only the left end-effector
is actuated toward a target cube. An episode succeeds when the active
end-effector comes within 0.05 m of the target; each episode runs up
to a fixed round budget (and is subject to an inherited 24\,s environment
auto-reset, \S3.5), where a \emph{round} is one VLM reasoning step
followed by an inverse-kinematics servo to the emitted target. All
evaluation runs use \texttt{--freeze\_level} (no curriculum auto-advance) so
every arm sees the identical task distribution.

\subsection*{3.2 Teacher and memory pipeline}

The teacher is Gemma-4-31B-it~\citep{gemma2024} (multimodal, 4-bit)
served via llama.cpp, augmented with an image-anchored episodic memory:
after each episode it writes a self-recap; at round 1 of a new episode it
retrieves the top-K visually-and-state-similar past recaps
(DINOv2~\citep{dinov2_2023} image embedding, state-aware ranking) and
injects them as a PAST\_LESSONS preamble: each retrieved episode's lesson
text and outcome, together with that episode's initial camera image.\footnote{\repopath{aiongenos/memory/retriever.py}.} The
memory-augmented teacher's trajectories are the distillation source;
its retrieval buffer, frozen at a fixed snapshot (SHA-pinned), is what
the C\_retrieval protocol later reuses at student inference. The recap
prompt received an unformatted \texttt{\{target\_color\}} instruction template
(disclosed in \S4.5); the lessons themselves are unaffected.

\subsection*{3.3 Students, arms, and the 2$\times$2 factorial}

Students are LoRA~\citep{lora2021} adapters (rank-16, $\alpha$=32, dropout 0.05) on the same
Gemma-4-31B-it base, targeting the language-model attention and MLP
projections only (\texttt{self\_attn.}\allowbreak\texttt{\{q,k,v,o\}\_proj},
\texttt{mlp.\{gate,up,down\}\_proj});
the vision tower is untouched. Each arm is trained in two stages on the
\emph{same} pinned trajectory pool: SFT (stage 4-A, 992 desirable
progress-rounds, 1 epoch, lr 2e-4) followed by composable-KTO~\citep{kto2024} refinement
(stage 4-B, 992 desirable + 1799 undesirable rounds, 1 epoch, lr 5e-5,
$\beta$=0.1, auto-balanced class weights) with the frozen SFT adapter as the
KTO reference (the C.3-B composable dual-adapter scheme). All arms share
this recipe and hyperparameter set; per-arm effective step counts are
logged for audit.\footnote{\repopath{training\_meta.json} (not released).}

The four training arms form a 2$\times$2 factorial over two binary factors:
retrieved-lesson gist in the target, and the teacher's native thought
in the target. All four terminate in the same canonical action lines,
synthesised from parsed coordinates by a single synthesiser across arms,
so output format is not a between-arm confound:

\begin{table}[h]
\centering
\begin{tabular}{lll}
\toprule
 & no gist & with gist \\
\midrule
\textbf{no native thought} & A\_action\_only & D\_gist \\
\textbf{with native thought} & A\_ctrl\_rat & B\_main \\
\bottomrule
\end{tabular}
\end{table}

All four arms are pinned to the identical 2791-round pool (992
desirable + 1799 undesirable), verified by SHA-256 before training
(Appendix A1, Amendments 12--13); this makes training-sample count a
controlled variable, not a confound.

\subsection*{3.4 Evaluation protocols}

Five inference protocols are evaluated, 100 episodes each, all at the
same task with \texttt{--freeze\_level}:

\begin{table}[h]
\centering
\begin{tabular}{lll}
\toprule
Protocol & Adapter & Inference-time retrieval \\
\midrule
A\_action\_only & own & none \\
A\_ctrl\_rat & own & none \\
B\_main & own & none \\
D\_gist & own & none \\
C\_retrieval & A\_ctrl\_rat's & frozen teacher buffer snapshot \\
\bottomrule
\end{tabular}
\end{table}

C\_retrieval is \emph{not} a fifth trained arm: it reuses A\_ctrl\_rat's
adapter weights and re-attaches the frozen retrieval buffer as an
inference-time PAST\_LESSONS preamble. Two contrasts therefore follow,
and we keep them distinct (a distinction the results section makes
central):
\begin{itemize}
\item \textbf{Identical-weights} (C\_retrieval vs A\_ctrl\_rat): same adapter,
  retrieval toggled. This is the clean parametric-vs-contextual contrast;
  it was not pre-registered as primary and is reported as exploratory.
\item \textbf{Registered protocol} (C\_retrieval vs B\_main = T4): each arm its
  own adapter, so this varies both adapter and protocol; pre-registered
  and confirmatory (Amendment 8).
\end{itemize}

The C\_retrieval buffer is read-only during evaluation; a tree-hash gate
confirms it was not written during the run (Amendment 12--13).

All five protocols share a fixed environment seed base (4500; per-
episode seed = 4500 + episode index), so episode \emph{k} starts from the
same initial world configuration across protocols, giving a paired design
(Amendment 11).

\subsection*{3.5 Pre-registered hypotheses and tests}

Registered before any adapter trained (pre-reg \S2--\S4, refined by
Amendments 9--10):

\begin{itemize}
\item \textbf{T1 (primary)}: B\_main $-$ A\_action\_only. Rejection: $\Delta$ SR $\ge$ +10 pp
  and significant, two-sided, $\alpha$=0.020. A graded reading within this
  budget line: T1-strong ($\ge$+10 pp) and T1-weak ($>$0, significant at
  $\alpha$=0.010). Explicit falsification clause on the memory-in-weights
  claim.
\item \textbf{T1a}: B\_main $-$ A\_ctrl\_rat, two-sided, $\alpha$=0.020.
\item \textbf{T4}: C\_retrieval $-$ B\_main, two-sided, $\alpha$=0.010.
\item \textbf{T3}: B\_main SR $\ge$ 0.7 $\times$ pooled memory-teacher SR (=36.2\%),
  one-sided.
\item Family-wise $\alpha$=0.060 (Amendment 10, non-uniform allocation).
\item \textbf{R1 $\Delta$X probe} (mechanism): round-1 target X minus initial EE X, per arm, in
  scalar-guard grid units (1 unit $=$ 0.45\,cm along x), against two reference
  fingerprints (memoryless-teacher $-$23.5, memory-teacher last-quartile $-$15.8),
  with a registered branch choice (H\_language vs H\_behavior). The per-arm
  prediction matrix is pre-registered (Amendment 9 \S9.3); the analysis lock
  lists the probe as exploratory (Amendment 14 \S14.5). The pre-registration
  labelled the units cm; the documentation-only units erratum (Amendment 16,
  \commit{730b9fc}) corrects this.
\item \textbf{Auto-reset sensitivity} (Amendment 15 pin 8, registered
  2026-10-05 before any Amendment 15 run): the task inherits a 24\,s
  environment time-out (720 steps) that auto-resets inside the environment
  step, which the orchestrator does not read, so long episodes continue on
  a re-sampled configuration (boundary at trajectory index 718, located per
  episode from the step counter). The sensitivity success rate counts only
  successes reached before that boundary. The as-run rate stays primary;
  the sensitivity value is reported beside every D11 success rate and
  contrast and never replaces it.
\end{itemize}

\subsection*{3.6 Analysis rules (locked after collection, before any p-value)}

Under the two-timestamp discipline (\S4.1), the analysis rules below were
locked (Amendment 14) after collection but before any p-value, with the
raw counts frozen out of the test-selection decision.

\begin{enumerate}
\item \textbf{Pairing-integrity gate}: for each episode index, a 5-way
   \texttt{allclose} ($\varepsilon$=1e-4 m) on the initial configuration; on failure, a
   mechanical fallback (below) applies, with no per-contrast discretion.
\item \textbf{Test selection}: if the gate passes, McNemar (paired) is primary; if it
   fails, the two-proportion z is primary. The choice is binary and all-or-nothing.
\item All pairwise tests two-sided (pre-reg \S4, never altered).
\end{enumerate}

In the event, the gate failed its literal check because replays do not
persist pre-action cube pose and the first trajectory sample is already
one servo-step in; a frozen-arm diagnostic (seed identical 100/100,
frozen right-arm identical 100/100, active left-arm 1/100) confirms the
pairing is physically real. Per the locked rule, the two-proportion z
governs; McNemar is reported as a sensitivity analysis and agrees on
all contrasts (Table~1).

\subsection*{3.7 Data curation history}

The training targets passed a deterministic three-rule rationale
filter (direction-consistency, ground-truth geometry, vacuity). Over
three blinded-audit rounds the first two rules were found to measure
\emph{coherence}, not \emph{correctness}, and were demoted to advisory flags.
Under the \texttt{flags\_only\_a6} policy actually used, the vacuity rule
also has zero drop authority (structurally guaranteed, Amendment 13), so
the filter drops no rows and the four arms' 992/2791 counts are
filter-invariant. The demotion is part of the data's provenance; the
full audit trail is in Appendix A2.

\section*{Appendix A1. Pre-registration amendment chain}

Every change to the frozen plan was filed as a signed, dated,
SHA-anchored amendment before the step it governed. Amendments 1--13
were filed before any adapter trained; Amendment 14 was filed after
collection but before any p-value.

\begin{table}[h]
\centering
\small
\begin{tabular}{@{}c l p{6.6cm} p{2.6cm}@{}}
\toprule
\# & Date & Change & Anchor \\
\midrule
1 & 07-06 & Sample-count control arm (B\_matched) added & pre-reg \S12 \\
2 & 07-06 & Filter + stratified blinded audit; escalation rule (agreement $<$85\% escalates to LLM judge) & --- \\
3 & 07-06 & Asymmetric KTO drop policy (desirable vs undesirable) + auto-balance invariant & \commit{2ba693c} \\
4 & 07-07 & Rule-1 sentence classifier (past/intent) fix & --- \\
5 & 07-07 & Rule-2 demoted to flag; composite-agreement rejected & \commit{6a29103} \\
6 & 07-07 & Coherence filter demoted to advisory flags (v-final 992+1810) & \commit{9b0a9da} \\
7 & 07-08 & Pre-flight checklist (length align, seed, env reset) & \commit{bc524b8} \\
8 & 07-08 & Cross-arm target hygiene + 2$\times$2 + C\_retrieval protocol & \commit{1a44d61} / \commit{a7bada2} \\
9 & 07-08 & T1 threshold, R1 probe branches, thought-leak, rename & \commit{503496b} \\
10 & 07-08 & B\_matched removed (confounder dissolved), $\alpha$ moved to T1a & \commit{bf2dbb5} \\
11 & 07-08 & Paired design with McNemar; env\_seed\_base=4500; SHA gate & \commit{bfa7cc6} \\
12 & 07-08 & Canonical filenames, D\_gist eval slot, SHA-gate timing & \commit{ae196b5} / \commit{0e26443} \\
13 & 07-08 & Step-2 row-count sentinel; structural \texttt{flags\_only\_a6} & \commit{9f5bec1} / \commit{cd1eb94} \\
14 & 07-13 & Analysis-time decision lock (gate, fallback, two-sided) & \commit{00be6d5} / \commit{73e712b} \\
\bottomrule
\end{tabular}
\end{table}

\emph{(Amendments 5--13 were filed 07-07/07-08, during the pre-flight period;
the anchors are the frozen commits, and the full text of each is in the
pre-registration document.)}

\emph{Counts are as pinned at each amendment's date; the final training
counts are 992 + 1799 (Amendment 8 re-pinned from A6's 992 + 1810 after a
trajectory-set ID diff removed 11 no-retrieval-hit rounds from a single
episode). Section 3 uses the final 1799.}

\section*{Appendix A2. Filter audit trail}

Three blinded audit rounds on the deterministic filter's drop surface;
Rules 1--2 precision on the correctness axis was low (they measure
coherence), motivating their demotion to flags (Amendments 4--6). The
\texttt{flags\_only\_a6} policy used in the final pipeline is a structural
zero-drop guarantee (Amendment 13): all three rules compute and emit
advisory flags, none drops a row, so the 992/2791 pool is
filter-invariant and the cross-arm SHA pins hold.

\section*{4. Results (Experiment 1)}

\subsection*{4.1 A pre-registered null: memory does not bake into weights (T1)}

Two timestamps matter and we keep them separate. The \textbf{hypothesis}
T1 was registered \emph{before any training} (pre-registration \S2, filed
2026-07-06; refined by Amendments 9--10 which sharpened the arm
semantics and reallocated $\alpha$). The \textbf{analysis rules} (pairing gate,
test-selection fallback, two-sided form) were locked \emph{after
collection but before any p-value was computed} (Amendment 14). The
raw success counts are visible in the collect logs; A14 froze only how
they would be tested, and the hypothesis predates it.

T1: a student distilled from memory-augmented-teacher trajectories
with retrieved-lesson gist in its target (\texttt{B\_main}) would exceed a
student distilled from the same trajectories with action-only targets
(\texttt{A\_action\_only}), evaluated with zero inference-time retrieval. The
pre-registered rejection region was $\Delta$ SR $\ge$ +10 pp \textbf{and} statistically
significant (two-sided, $\alpha$=0.020), with an explicit falsification clause.

This contrast is confirmatory. \texttt{B\_main} scored 26/100 and \texttt{A\_action\_only} 25/100, with no
detectable effect: +1 pp (z=0.16, p=0.87, two-sided; Newcombe 95\% CI
[$-$11.0, +13.0]; McNemar sensitivity p=1.0, agreeing). Auto-reset
sensitivity (\S3.5): 19 vs 21, $-$2 pp (z=$-$0.35, p=0.72), same verdict.

We are deliberate about what this licenses. The registered minimum
effect (+10 pp) lies \emph{inside} the confidence interval, and a post-hoc
minimum-detectable-effect calculation is unforgiving: at n=100 per arm
with SR$\approx$0.25, significance at the registered two-sided $\alpha$=0.020 requires
$\Delta$ $\ge$ 14.2 pp, so an \emph{observed} +10 pp difference could not reach
significance, and the design has only about one-in-four power for its own
registered +10 pp effect. This is a flaw in our registered design, and we
disclose it. Accordingly, \textbf{per the registered clause we
withdraw the memory-in-weights claim (we stop asserting it), but we do
not assert the strong null at the success-rate level}, which would read
absence of evidence as evidence of absence. The claim that the
\emph{conditional} structure specifically fails to transfer does not rest on
this underpowered SR test; it is carried by the R1 mechanism
probe (\S4.3) and, at round 1, by the residual diagnostic (\S4.4b), whose
round-1 null is not subject to the SR power problem (with its own
$|r|<0.20$ caveat; later rounds are confounded, \S4.4b).

Scope: this null concerns \emph{this distillation recipe}, namely single-round
SFT + composable-KTO on the L0a-Left reach task, at this data scale
and adapter capacity (LoRA rank-16). \S4.4b and \S4.5 sharpen the data-
and capacity-scope respectively. It does \textbf{not} claim that baking-in is
impossible. Stronger consolidation (multi-round, replay, surprise-gated
writing) is untested and named as future work (\S6).

\subsection*{4.2 The reversal: retrieval on fixed weights nearly triples success}

\begin{figure}[htbp]
  \centering
  \includegraphics[width=\linewidth]{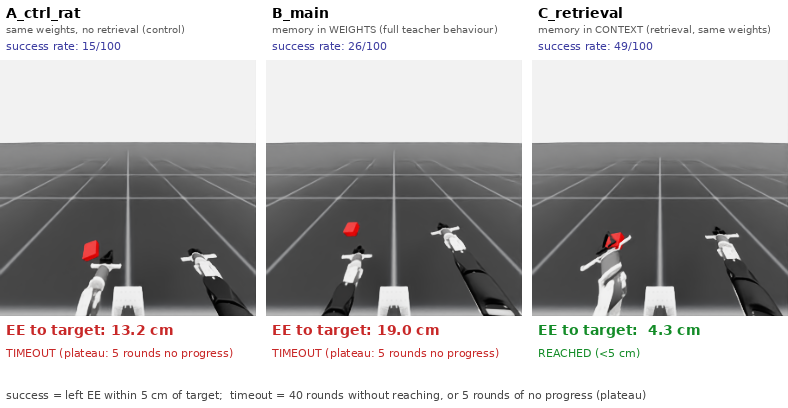}\\[2pt]
  \includegraphics[width=0.72\linewidth]{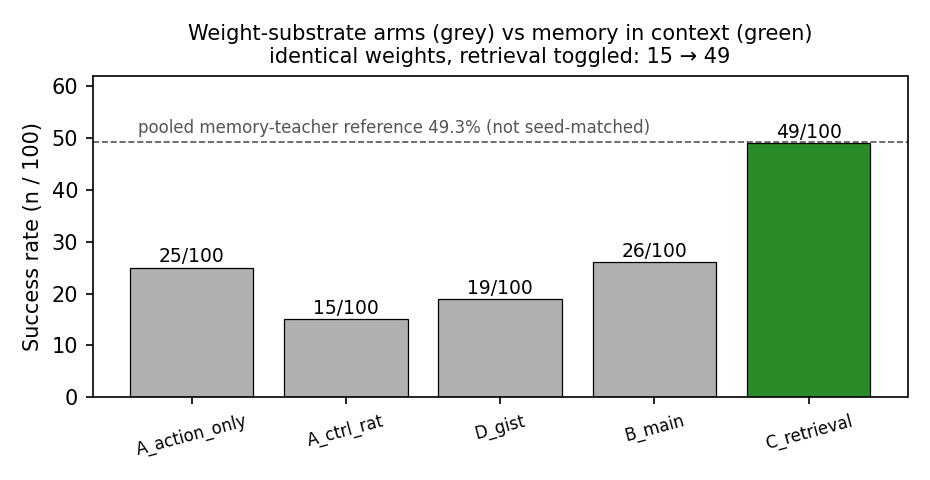}
  \caption{\textbf{Where memory lives decides whether it transfers.}
  \emph{Top:} one representative episode per arm (final frame; the red cube
  is the left-arm target). \texttt{A\_ctrl\_rat} (the same-weights,
  no-retrieval control) and \texttt{B\_main} (the full-teacher-behaviour
  distilled arm) both time out, while \texttt{C\_retrieval} (the
  \emph{same} \texttt{A\_ctrl\_rat} weights with an inference-time retrieval
  preamble) reaches the target. Overlaid is the live
  end-effector-to-target distance; each panel's success rate ($n/100$) is
  the official episode-level rate, so a single rollout is one representative
  sample of 100, not the evidence itself. \emph{Episode selection:} the
  early collect did not persist the environment seed, so the three panels
  are not the same seed; each is the episode nearest its arm's median
  outcome (failures near the median failure, not extremes), with matched
  initial EE-to-target distance ($\approx$20\,cm; 19.2--20.3\,cm for the three shown). \emph{Termination criteria
  (identical for all arms):} a \emph{round} is one VLM step + IK servo;
  \textbf{success}: the scored (left) arm is within $0.05$\,m of target;
  \textbf{timeout}: either 40 rounds elapse without reaching, or a plateau
  (5 consecutive rounds of no mean progress). \emph{Bottom:} official
  success rate for all five arms. The four weight-substrate arms (grey)
  all sit near baseline; memory-in-context (\texttt{C\_retrieval}, green)
  reaches the memory-teacher's own 49.3\% (dashed). An animated version is
  in the supplementary material.}
  \label{fig:qual}
\end{figure}

Two contrasts speak to bake-in versus retrieval (Figure~\ref{fig:qual}); we
report both with their exact adapter relationships.

\textbf{The identical-weights contrast (exploratory, decisive).} The genuine
same-weights comparison is \texttt{C\_retrieval} vs \texttt{A\_ctrl\_rat}: \texttt{C\_retrieval}
\emph{is} \texttt{A\_ctrl\_rat}'s adapter, run with an inference-time frozen-buffer
retrieval preamble; \texttt{A\_ctrl\_rat} is the same adapter with none. On
\emph{fixed} student weights, attaching retrieval moved success from 15/100 to 49/100
(auto-reset sensitivity 12 to 48, +36 pp, z=5.56), a difference of
\textbf{+34 pp (z=5.15, p$\approx$2.6$\times$10$^{-7}$; Newcombe 95\% CI [+21, +45])}. This was
not pre-registered as a primary test, so we mark it exploratory. At
p$\approx$3$\times$10$^{-7}$ it survives any multiplicity correction; exploratory here
means ``not registered'', not ``fragile''.

\textbf{The registered protocol contrast (confirmatory).} The
pre-registered T4 is \texttt{C\_retrieval} vs \texttt{B\_main}, a \emph{protocol} contrast
in which each arm carries its own adapter (C\_retrieval = A\_ctrl\_rat's
weights + retrieval; B\_main = the gist+thought-baked weights, no
retrieval). The confirmatory result is 49/100 vs 26/100, a difference of +23 pp (z=3.36,
p=7.8$\times$10$^{-4}$; McNemar p=1.0$\times$10$^{-3}$, agreeing), significant at the
pre-registered $\alpha$=0.010 (auto-reset sensitivity 48 vs 19, +29 pp, z=4.35). This contrast varies both adapter and protocol,
so it under-states the pure retrieval effect the identical-weights
contrast isolates.

In one line: \emph{on fixed student weights, attaching retrieval
added +34 pp (exploratory, z=5.15); the registered
consolidation-vs-retrieval protocol contrast was +23 pp (confirmatory,
z=3.36).} Both point the same way: memory supplied through context at
inference vastly outperforms the same memory routed through
distillation into weights. The pre-registration \S4 had anticipated a
\emph{small} C\_retrieval$-$B\_main gap (``bake-in replaces retrieval''); the
\textbf{interpretation} inverts, while the test itself is unchanged.

\emph{Registration note.} A reader may object that the direction was seen
first. The two-sided form of every pairwise test was
registered pre-collection: \S4 of the original pre-registration (dated
2026-07-06) reads verbatim ``all two-proportion z, $\alpha$=0.05, \textbf{two-sided}''.
The significance threshold for T4 ($\alpha$=0.010) was set in Amendment 1
(restated in Amendment 11), also before collection. A14 restated the two-sided form; it did not
introduce it. T4 is thus a pre-registered, two-sided confirmatory
result whose rejection was reachable in either direction, so the test
design did not privilege the +23 pp outcome.

One further disclosure concerns the A14 lock. The pairing-integrity gate's
\emph{literal} failure was foreseeable at lock time, because replays not
persisting the pre-action cube pose is a fixed schema property known in
advance. The fallback from the gate to the z-test
was therefore a mechanical rule, not a branch chosen after seeing which test
favoured which conclusion. We therefore lean on the fact that the two
tests \emph{agree on every contrast} (McNemar and z, Table~1), rather than on a claim
that the fallback was executed blind. On the one contrast where the
branch could matter (T1a: McNemar p=0.046 vs z p=0.054), both remain
n.s. at the pre-registered $\alpha$=0.020, so the choice changes no verdict.

\textbf{The registered transfer floor (T3), and the like-for-like reading
it forces.} This test is confirmatory. T3 asks whether \texttt{B\_main} (distilled,
no-retrieval) clears 0.7 $\times$ the \emph{memory-augmented} teacher's 51.7\% pool
= 36.2\%. It does not: 26\% $<$ 36.2\% (one-sided z=$-$2.12; sensitivity 19\%, z=$-$3.58). T3
\textbf{fails}. The floor, however, is a cross-rung comparison: it
measures a \emph{no-memory} student against a \emph{with-memory} teacher's bar,
a level mismatch built into the registration. The like-for-like alignment
of rungs below is exploratory.

\begin{table}[h]
\centering
\small
\begin{tabular}{l l l}
\toprule
rung & student & teacher (same rung) \\
\midrule
no-memory & B\_main / A\_action\_only $\approx$ 25--26\% (sens.\ 19--21) & memoryless teacher 22\%$\dagger$ (sens.\ 19) \\
with-memory & C\_retrieval 49\% (sens.\ 48) & memory-teacher 49.3\%$\ddagger$ (sens.\ 47.5) \\
\bottomrule
\end{tabular}
\end{table}

At \emph{matched} memory conditions the student tracks the teacher on both
rungs; what fails is specifically the middle step, \textbf{consolidating the
memory advantage into weights}. The practitioner take-home is therefore not
``competence didn't transfer'' (it did, at each rung) but the sharper
recipe: \textbf{externalise the memory}. A
student carrying a frozen retrieval buffer reaches the memory-teacher's
own rung (49\% vs 49.3\%; sensitivity 48\% vs 47.5\%), whereas routing that same memory through distillation
into weights lands it back on the no-memory rung. The competence half of
the recipe (``distil the competence'') does not survive the Amendment 15 controls (\S4.2b):
with the buffer attached, the base model with no adapter does as well
(48 vs 49 of 100; sensitivity 48 vs 48).

$\dagger$Memoryless teacher = D6b, the teacher with memory off but the
same Fix 1/3 scaffolding as the memory teacher, measured directly: 22/100
(run \texttt{fa7f4571}). Like the memory-teacher pool it is not
seed-matched to the D11 eval.
$\ddagger$The two teacher figures use different pools and we keep them distinct
on purpose: T3's floor uses the pre-registration's pinned 51.7\%
(pre-reg \S4, the memory-teacher pool at registration time); the parity
reference uses the post-fix pool's 49.3\% (n=221). Neither is
seed-matched to the D11 eval, so both are level-setting references, not
controlled contrasts; we report both rather than picking one.

\FloatBarrier
\subsection*{4.2b Adapter controls: the buffer without the adapter, and on the other adapters (Amendment 15)}

Amendment 15 (lock \commit{989a753}, 2026-10-05) was registered after the
D11 results and before any of its episodes ran. It attaches the same frozen
buffer, read-only (tree hash unchanged after every run), to (a) the base
model with no adapter and (b) each of the three adapters
\texttt{C\_retrieval} did not use, on the same 100 seeded configurations.
Seeds match in every contrast (0/100 mismatches), but the first-step
fingerprint gate fails by instrument design, as in D11, so the
two-proportion $z$ is primary by the Amendment 14 \S14.2 fallback; the
amendment fixes no $\alpha$, and we use the \S4 default of 0.05, two-sided.
Table~\ref{tab:a15} gives the counts.

\begin{table}[h]
\centering
\caption{Amendment 15 adapter controls ($n$=100 per arm; auto-reset sensitivity counts in parentheses). $\Delta$ against the same weights' D11 run without retrieval, and against \texttt{C\_retrieval}; two-proportion $z$, two-sided. R1$'$ slope: Spearman $r$ of round-1 $\Delta$X against the required correction under the 15\,mm recovery gate (A15.1; pre-registered reading for the four with-buffer rows).}
\label{tab:a15}
\resizebox{\textwidth}{!}{%
\begin{tabular}{l l l l l l}
\toprule
Weights + frozen buffer & SR & without buffer (D11) & $\Delta$ vs own, $z$ & $\Delta$ vs C\_retrieval, $p$ & R1$'$ $r$ (above band) \\
\midrule
base, no adapter (a) & 48 (48) & --- & --- & $-$1 pp, 0.887 & +0.29 (yes) \\
A\_action\_only (b) & 53 (53) & 25 (21) & +28 pp, 4.06 & +4 pp, 0.572 & +0.49 (yes) \\
B\_main (b) & 53 (50) & 26 (19) & +27 pp, 3.91 & +4 pp, 0.572 & +0.36 (yes) \\
D\_gist (b) & 55 (55) & 19 (17) & +36 pp, 5.27 & +6 pp, 0.396 & +0.44 (yes) \\
\bottomrule
\end{tabular}%
}
\end{table}

\textbf{(a) The adapter is not needed under retrieval.} This test is confirmatory for the Amendment 15 prediction ($\alpha$ = 0.05 is a disclosed choice), and the registered prediction was not met. The
prediction was that the base model with the buffer would fall
below \texttt{C\_retrieval}, the gap measuring the adapter's contribution.
It scored 48 against 49: $-1$ pp, 95\% CI $[-14.6, +12.6]$, $z{=}{-}0.14$,
$p{=}0.887$, inside the paired MDE (19.8 pp). The interval does not exclude
an adapter contribution of up to 14.6 pp, so this is an undetected
contribution at $n{=}100$, not a demonstrated zero. The locked fallback
applies as written: ``distil the competence'' is downgraded, and the recipe
is rewritten to externalise the memory.

\textbf{(b) The effect does not depend on the adapter.} This test is confirmatory on the same terms, and the registered prediction was met for all three adapters (3/3). The
prediction was that all three adapters would rise into the
teacher band (45--55\%). They did: from 25, 26 and 19 to 53, 53 and 55
(sensitivity: from 21, 19 and 17 to 53, 50 and 55; all $p<10^{-4}$), and none differs from \texttt{C\_retrieval}
($|\Delta|\le 6$ pp, $p\ge 0.39$). The auto-reset sensitivity counts change
no verdict. With the buffer, success is 48--55 whatever weights are loaded
(sensitivity 48--55); without it, 15--26 (sensitivity 12--21). The $+34$ pp of \S4.2 is therefore not specific to
\texttt{A\_ctrl\_rat}'s weights.

\textbf{R1$'$ slope (A15.1).} Every with-buffer arm's round-1 move tracks
the required correction ($r$ above its permutation band, $+0.29$ to $+0.49$,
no episode excluded by the gate). For context only, and post hoc: the D11
\texttt{C\_retrieval} arm gives $r{=}{+}0.31$, and none of the four adapters
without the buffer is above its band ($r{=}{-}0.30$ to $-0.16$). This is the
conditional-correction signature of \S4.3, now read as a slope rather than
a $\sigma$. As everywhere in this study, the retrieved context carries the
lesson text and its episode image together, so the two are not separated.

\FloatBarrier
\subsection*{4.3 Mechanism: distillation bakes the margin, retrieval supplies the condition (R1 $\Delta$X probe)}

The R1 $\Delta$X behavioural probe (round-1 target X minus initial EE X)
points to \emph{what} transferred (exploratory). Its reference fingerprints
are the D6 memoryless teacher ($-$23.5) and the memory-teacher last
quartile ($-$15.8). All R1 values are in scalar-guard grid units: the
pre-registration labelled them ``cm'', but the probe takes the difference
of two integer-grid coordinates, and 1 unit $=$ 0.45\,cm along x (units
erratum, Amendment 16, \commit{730b9fc}, documentation-only). This probe was
\textbf{pre-registered with a per-arm prediction matrix} (Amendment 9 \S9.3)
that named two branches: an H\_language branch (bias correction carried
by rationale text, so only gist/thought arms would shift) and an
H\_behavior branch (correction carried by the action distribution, so all
mem-teacher-distilled arms shift regardless of rationale).

This result is exploratory: the per-arm predictions were pre-registered,
and the analysis lock (Amendment 14 \S14.5) lists the probe as
exploratory. All four distilled arms cluster at the
memory-teacher fingerprint rather than D6's, and the control \texttt{A\_action\_only}
(never trained on any rationale) shifts identically. As an exploratory
reading, this selects the \textbf{H\_behavior branch}: the carrier of the
bias correction is the action distribution, not the memory content.
Round-1 $\Delta$X uses indices 0--29,
far before the auto-reset boundary, so the sensitivity count does not
apply.

\begin{table}[h]
\centering
\begin{tabular}{l l l l}
\toprule
Arm & R1 $\Delta$X $\mu$ & $\sigma$ & closer to \\
\midrule
A\_action\_only & $-$16.62 & 8.61 & mem-teacher ($-$15.8) \\
A\_ctrl\_rat & $-$15.75 & 10.11 & mem-teacher \\
B\_main & $-$16.67 & 8.53 & mem-teacher \\
D\_gist & $-$18.15 & 7.89 & mem-teacher \\
\textbf{C\_retrieval} & \textbf{$-$7.10} & \textbf{14.39} & (bias nearly erased) \\
\bottomrule
\end{tabular}
\end{table}

All four arms were pinned
(Amendment 8) to the same memory-teacher trajectories, whose coordinates
were already memory-corrected; distillation moved the corrected
behavioural \emph{average} into weights, and all four arms inherited it
equally, which is exactly why T1 is +1 pp.

\textbf{Variance analysis (exploratory).} What
distillation does \textbf{not} transfer is visible in the variance.
\texttt{C\_retrieval}'s $\mu$=$-$7.10 with $\sigma$=14.4 (the largest) is the signature of a
\emph{conditional} correction: retrieval applies a different, situation-
specific adjustment per episode, averaging near zero bias with wide
spread. The four distilled arms apply the \emph{same} static $-$16.7 shift to
every episode regardless of need. This motivates the mechanism reading:
\emph{behavioural} distillation of a \emph{prompt-retrieved} memory teacher bakes
in the marginal distribution shift (a static prior) but not the
conditional structure (situation-specific correction). The conditional
correction is carried by the retrieved context (lesson text together with
its episode image); the present design does not separate the two
components. The conditional component does not survive this recipe and is
the source of retrieval's edge. We scope the
claim to prompt-retrieved memory and behavioural distillation
deliberately: a teacher whose memory is architecturally trained-in
(\S6.2) presents a different distillation surface we do not test. This
variance-based account is
exploratory: it interprets $\sigma$, which the pre-registration did not
predict, whereas the $\mu$-based branch selection above tests
pre-registered per-arm predictions (itself exploratory under the analysis
lock, Amendment 14 \S14.5).

\begin{figure}[htbp]
  \centering
  \includegraphics[width=0.82\linewidth]{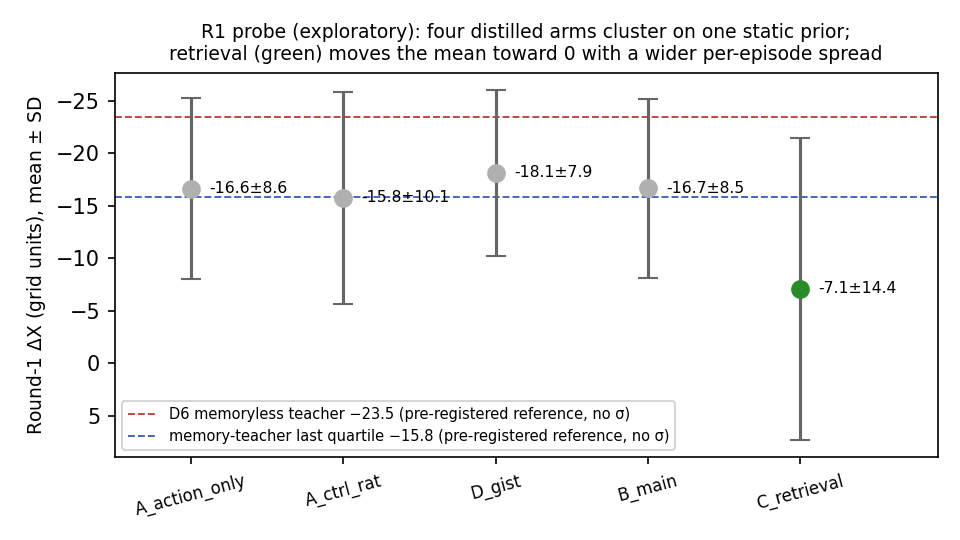}
  \caption{\textbf{The R1 mechanism, one panel.} Round-1 $\Delta$X toward
  the target (scalar-guard grid units; 1 unit $=$ 0.45\,cm along x), per arm, against two registered reference fingerprints:
  the D6 memoryless teacher ($-$23.5) and the memory-teacher last quartile
  ($-$15.8). The four distilled arms sit tight on the $-$15.8 line at
  $\approx$$-$16.7$\pm$9, one static prior every arm adopts regardless of
  episode, while \texttt{C\_retrieval} shifts to $-$7.1$\pm$14.4: mean
  toward zero \emph{and} a wider, per-episode spread, the signature of a
  conditional correction. Mean $\pm$ SD per arm, plotted from the per-arm
  summary statistics stored by the original exploratory output; per-episode
  values are recomputable from the replays. The $\sigma$ contrast, $8$--$10$ vs
  $14.4$, is the reported effect.}
  \label{fig:r1}
\end{figure}

The R1 signature (Figure~\ref{fig:r1}) makes the split visible in one
panel: a static prior (tight, on the memory-teacher line) for the
distilled arms, a conditional correction (mean toward zero, wide spread)
for retrieval.

\FloatBarrier
\subsection*{4.4 Which episodes need memory, and what ``conditional'' is not}

The analyses in this subsection are exploratory. Exploiting the shared-seed
paired design, 29/100 episodes are solved \emph{only} by \texttt{C\_retrieval} (all
four no-retrieval arms fail); we call this the ``memory-dependent'' episode set.

Two exploratory mechanistic hypotheses for \emph{what makes them
memory-dependent}, each stated in the audit log before its test,
returned null:
\begin{itemize}
\item \textbf{Not starting difficulty}: rescue-only episodes are not init-pose
  outliers (init-pose deviation ratio 0.96; initial red-cube distance
  identical, 19.9 cm vs 19.9 cm). Result: null.
\item \textbf{Not mid-trajectory recovery}: \texttt{C\_retrieval} shows a non-monotonic
  diverge-then-converge signature in only 10\% of rescue episodes (other
  arms show it \emph{more}, 22\%, on the same episodes). Its rescues are
  monotonic convergence from the first round. Result: null.
\end{itemize}

Both nulls narrow the mechanism rather than weaken it. Retrieval's edge
is not any single timepoint (hard start, or a mid-course save) but
\textbf{per-round sustained supply of the correct direction}: it emits a
good target every round, whereas the baked-in static prior gives one
round-1 correction and no per-round situational update. Failing arms
stall for lack of continued correct supply; recovery is not what they lack.
A round-resolved test of this is future work (\S6).

The rescue structure quantifies this: retrieval solves 44 episodes that
\texttt{A\_ctrl\_rat} fails and 34 that \texttt{B\_main} fails, with 29
episodes solved \emph{only} when memory is supplied in context (the
pure-memory-dependent set). We report these counts in text rather than as
a separate UpSet figure; the per-episode membership lists are in the
exploratory audit output.\footnote{\repopath{exploratory.json}, field \texttt{rescue} (not released).}

\subsection*{4.4b Is the null just too little data? A no-training diagnostic}

The first reviewer question about a null this size is whether \texttt{B\_main}
underperformed for lack of training data: 992 desirable rounds spread
over a continuous cube-position space may leave each situation-cell
sparse, and a conditional map from situation to correction is far more
sample-hungry than a constant marginal shift. The concern is
well-founded. The marginal/conditional asymmetry has a sample-complexity
asymmetry built in, and this is arguably the \emph{cause} of our headline
split rather than a threat to it.

We can bound it without any retraining, using the paired seeds, in an
exploratory diagnostic. We treat \texttt{C\_retrieval}'s per-episode R1 residual (its
round-1 correction minus its own mean) as a proxy for the
situation-specific correction each episode calls for, and correlate it
with each distilled arm's R1 residual on the same episode. If weight
distillation captured any conditional structure, the correlation should
be positive; a null correlation means the arm learned nothing beyond
the static prior.

\begin{table}[h]
\centering
\begin{tabular}{l l l l}
\toprule
distilled arm & r (round 1) & r (round 2) & r (round 3) \\
\midrule
A\_action\_only & $-$0.12 (0.23) & +0.05 (0.70) & +0.18 (0.088) \\
B\_main & $-$0.12 (0.22) & +0.05 (0.65) & +0.40 (0.001) \\
A\_ctrl\_rat & +0.03 (0.78) & +0.11 (0.30) & +0.27 (0.013) \\
D\_gist & $-$0.12 (0.23) & +0.10 (0.34) & +0.12 (0.27) \\
\midrule
n (paired episodes) & 100 & 89--90 & 84--85 \\
\bottomrule
\end{tabular}
\caption{Correlation of each distilled arm's per-episode residual with
\texttt{C\_retrieval}'s, by round (two-sided permutation p, 2000 draws).
All rounds are a \emph{2026-10 post-hoc recomputation} from the persisted
replays with the round-1 definition; round 1 reproduces the original
exploratory output exactly.}
\label{tab:residual_rounds}
\end{table}

\textbf{Round 1.} Every round-1 correlation falls inside the chance band
($\pm 1.96/\sqrt{n} \approx \pm$0.20 at n=100); none is significant, and
\texttt{B\_main} does not exceed the no-rationale control \texttt{A\_action\_only}.
Two caveats bound the reading. (i) \texttt{C\_retrieval}'s correction is one
noisy sample of a good policy, not ground truth, so the correlation is
attenuated toward zero. (ii) With n=100 the correlation's 95\% CI is
roughly $\pm$0.20, so this rules out a \emph{moderate} round-1 conditional
structure, not a vanishingly weak trace. The near-uniform $-$0.12 is
plausibly a shared-residual artifact of the common seed/trajectory source;
it lies inside the chance band.

\textbf{Rounds 2--3 (exploratory, 2026-10 post-hoc).} At round 2 all
correlations are small and non-significant. At round 3 every distilled arm
correlates positively with \texttt{C\_retrieval}'s residual (+0.12 to +0.40;
permutation-significant for \texttt{B\_main}, +0.40, p=0.001, and
\texttt{A\_ctrl\_rat}, +0.27, p=0.013). This is consistent with fresh visual
observations at later rounds driving situation-dependent behaviour, but two
confounds prevent reading it as a distilled conditional trace.
\emph{Shared target geometry:} paired episodes share the cube, and the
distilled arms correlate with \emph{each other} at round 3 by +0.12 to
+0.32 with no retrieval involved. \emph{Survivorship:} successful episodes
end early, leaving n$\approx$85 at round 3. Nor does gist-in-target separate
from the control here: \texttt{B\_main}'s +0.40 vs \texttt{A\_action\_only}'s
+0.18 is not a significant difference (Fisher z=1.53, p=0.13), and we do not
claim that \texttt{B\_main} exceeds the control.

Accordingly the claim is scoped to round 1: \textbf{at round 1 we find no
detectable conditional trace} ($|r| < 0.20$), and at round 1
\texttt{B\_main} $\approx$ \texttt{A\_action\_only} mirrors T1a at finer grain:
gist-in-target contributes no detectable round-1 conditionality. Whether
more data of the same kind would grow a conditional component is
\textbf{open, not excluded}: the later-round signal is confounded rather
than absent, so this diagnostic does not license skipping a data-scaling
probe. We did not run one; it is the first data-side follow-up, alongside
the architectural route of \S6.2.

\subsection*{4.5 Limitations}

\begin{itemize}
\item \textbf{Rationale-fabrication is a measurement gap, not a null.} The D11
  collect did not persist raw VLM text (the constrained-decode path
  stored only parsed coordinates), so per-arm rationale-fabrication rate
  and self-produced-gist content are \emph{not computable} from these runs.
  We do not substitute a proxy. The replay
  schema now stores reset-time pose + seed; paper 2's re-run adds
  raw-response logging.
\item \textbf{Pairing gate literal failure.} The confirmatory analysis used the
  pre-registered mechanical fallback (two-proportion z) because the
  pairing-integrity gate failed its literal check: replay does not
  persist pre-action cube pose and trajectory[0] is already one
  servo-step in. A frozen-arm diagnostic (seed identical 100/100,
  frozen right-arm identical 100/100, active left-arm 1/100) shows the
  pairing is physically real; the reported McNemar sensitivity
  (agreeing with z on all contrasts) recovers the design's power. Per
  the pre-registration, the z verdict governs.
\item \textbf{Adapter capacity is a sibling explanation to data quantity.} A
  rank-16 LoRA may lack the capacity to represent a conditional
  function from situation to correction even with unlimited data. This limitation is
  parallel to, not subsumed by, the data-scaling diagnostic in \S4.4b
  (that probe holds capacity fixed and varies data; this holds data
  fixed and varies capacity). We do not run a rank sweep here. Together
  these sharpen the scope of the T1 null to: \emph{this recipe, at this data
  scale and this adapter capacity}.
\item \textbf{No inference-cost advantage (measured; descriptive, non-inferential).} A same-hardware
  re-measure (Amendment 15 add-on; 20 steps per protocol on the same
  llama-server)\footnote{\repopath{logs/cost\_remeasure\_20261005\_181930.md} (not released).} gives:

  \begin{center}
  \begin{tabular}{l r r}
  \toprule
  protocol & completion tokens / step & wall-clock s / step \\
  \midrule
  memory teacher & 207 & 14.1 \\
  student (\texttt{A\_ctrl\_rat} weights) & 186 & 11.2 \\
  student + retrieval (\texttt{C\_retrieval}) & 220 & 13.7 \\
  \bottomrule
  \end{tabular}
  \end{center}

  The three are of the same order, and we make no inference-cost claim.
  Teacher and student share the 31B base, so per-token decode speed is
  the same (44--48 ms/token), and decode accounts for 70--73\% of each
  step's wall-clock. The student generates as much as the teacher because
  its training target carries a rationale before the subgoal, as the
  teacher's output carries one before its action. A latency advantage
  would need a smaller student or a learned action decoder (\S6.1).
\item \textbf{Two instrument defects (disclosed; Amendment 15 P1 pack).}
  (i) The task inherits a 24\,s environment time-out that auto-resets
  inside the environment step; the orchestrator does not read the
  truncation flag, so episodes longer than 718 steps continue on a
  re-sampled configuration. The pre-registered sensitivity count
  (\S3.5) is reported beside every D11 success rate and contrast; it
  changes no verdict. (ii) The recap prompt received the unformatted
  \texttt{\{target\_color\}} template: the literal appears in the
  instruction field of all 547 frozen-buffer entries and in none of the
  lessons (517/547 lessons name the colour); retrieval injects lessons
  and their episodes' initial images, not the instruction. The instrument is frozen and not patched.
\item \textbf{Text and image are not separated.} Each retrieved episode
  enters the prompt as its lesson text together with its initial image
  (\S3.2). Text and image contributions to the retrieval effect are not
  separated in this study, and the two routes deliver the buffer in
  different forms.
\item \textbf{The R1$'$ recovery gate was re-set after the pinned one
  failed.} Amendment 15 pinned a 1\,mm gate before the residual had been
  measured. On D11 data it excluded most episodes for an instrument reason
  (the first trajectory sample is logged after the first execute step), so
  A15.1 re-registered it at 15\,mm, above the 12.1\,mm maximum residual
  over the D11 arms. This was done before any Amendment 15 slope was
  computed, but after the D11 slopes at 15\,mm had been seen. The
  Amendment 15 arms then reached residuals of up to 13.4\,mm, 1.6\,mm under
  the gate. None was excluded, though the margin is thin.
\item \textbf{Single task, single recipe.} L0a-Left only; one distillation
  configuration. Generalisation across tasks and stronger consolidation
  recipes is untested.
\end{itemize}

\begin{table}[h]
\centering
\caption{Confirmatory contrasts ($n$=100/arm, two-sided $z$ primary).}
\label{tab:confirmatory}
\resizebox{\textwidth}{!}{%
\begin{tabular}{l l l l l l l l l}
\toprule
Test & Contrast & $\Delta$ SR & $\Delta$ 95\% CI (Newcombe) & z & p\_z & McNemar p & agree & verdict @ pre-reg $\alpha$ \\
\midrule
T1 & B\_main $-$ A\_action\_only & +1 pp & [$-$11.0, +13.0] & 0.16 & 0.87 & 1.0 & \checkmark & FAIL @ $\alpha$=0.020 $\rightarrow$ \textbf{null} \\
T1a & B\_main $-$ A\_ctrl\_rat & +11 pp & [$-$0.2, +22.0] & 1.93 & 0.054 & 0.046 & \checkmark & \textbf{n.s.} @ $\alpha$=0.020 \\
T4 & C\_retrieval $-$ B\_main & +23 pp & [+9.6, +35.3] & 3.36 & 7.8e-4 & 1.0e-3 & \checkmark & \textbf{PASS} @ $\alpha$=0.010 \\
T3 & B\_main SR vs 0.7$\times$teacher floor & 26\% vs 36.2\% & --- & $-$2.12 & 0.98$^{1}$ & --- & --- & \textbf{BELOW floor} \\
\midrule
\multicolumn{9}{l}{\emph{Auto-reset sensitivity (Amendment 15 pin 8; reported beside, never replacing, the rows above)}} \\
T1 sens. & 19 vs 21 & $-$2 pp & [$-$13.1, +9.1] & $-$0.35 & 0.72 & 0.80 & \checkmark & null \\
T1a sens. & 19 vs 12 & +7 pp & [$-$3.2, +17.1] & 1.37 & 0.17 & 0.14 & \checkmark & n.s. \\
T4 sens. & 48 vs 19 & +29 pp & [+16.0, +40.7] & 4.35 & 1.4e-5 & 4.4e-5 & \checkmark & PASS \\
T3 sens. & 19\% vs 36.2\% & --- & --- & $-$3.58 & 1.00 & --- & --- & below floor \\
\bottomrule
\end{tabular}%
}

\smallskip
\begin{minipage}{\textwidth}
\footnotesize
$^{1}$ T3 is a one-sided proportion-vs-constant test (H$_1$: SR $>$ floor); floor
= 0.7 $\times$ 51.7\% pooled memory-teacher SR = 36.2\% (pre-reg \S4). Reported
one-sided p is the upper-tail probability; B\_main is below the floor.

\smallskip
\textbf{Two-tier T1}: T1-strong ($\Delta$ $\ge$ +10 pp \& sig) and T1-weak ($\Delta$ $>$ 0 \& sig
at $\alpha$=0.010) are graded readings \emph{within} T1's single $\alpha$=0.020 budget
line (Amendment 9 \S9.2), not separate family members. Both fail here.

\smallskip
\textbf{$\alpha$ budget (Amendment 10 \S10.2, family-wise 0.060)}: T1 0.020, T1a
0.020, T3 0.010, T4 0.010. (The original \S4 used $\alpha$=0.05 per test; Amendment 1 set a
five-test non-uniform allocation summing to 0.060; Amendment 10 dropped
B\_matched and moved its 0.010 to T1a, keeping 0.060. The documented
over-spend buys power on the T1a safety-net contrast and is itself
pre-registered.)

\smallskip
\textbf{Reading the CIs}: T1a's [$-$0.2, +22.0] states that
a real effect may exist but this study is underpowered to resolve it
(the interval barely includes zero); we therefore report n.s. at the
pre-registered $\alpha$ and make no directional claim. T4's [+9.6, +35.3]
excludes zero comfortably.
\end{minipage}
\end{table}

\begin{table}[h]
\centering
\caption{Arm / protocol definitions (from pre-reg \S10.3).}
\label{tab:arms}
\begin{tabular}{l l l l}
\toprule
Protocol & Adapter & Training target & Inference retrieval \\
\midrule
A\_action\_only & own & canonical action only & none \\
A\_ctrl\_rat & own & native thought + action & none \\
B\_main & own & gist + native thought + action & none \\
D\_gist & own & gist + action & none \\
C\_retrieval & A\_ctrl\_rat's & (reuses A\_ctrl\_rat) & frozen D10-ext buffer \\
\bottomrule
\end{tabular}
\end{table}

The L2 design is fixed by the L2 pre-registration (Amendments 1/1a/2/3),
all frozen before any L2 eval number; as in the D11 Methods, every choice
below traces to a dated, numbered amendment.

L2 reuses the D11 apparatus unchanged (teacher/memory pipeline, LoRA
recipe, seed-paired design, McNemar/z machinery, \texttt{flags\_only\_a6}
filter). Below we specify only what is new for
L2; references to ``\S3.x'' point to the
Experiment-1 Methods.

\section*{5. Experiment 2 (L2 position-goal reach)}

\subsection*{5.1 Task: dual-arm position-goal reaching in a 6-DoF control space (Amendment 1 \S2)}

L2 is a \textbf{dual-arm position-goal reaching task in a 6-DoF control space}
(working title ``dual\_push'', renamed in Amendment 1 \S2). Both
end-effectors are driven toward per-arm goals (\texttt{left\_ee\_pose} /
\texttt{right\_ee\_pose} commands tracked under
\texttt{end\_effector\_position\_tracking} rewards); the goal cubes are
target visualizers, not pushable objects.\footnote{Task inspection: \repopath{docs/paper/l2\_diagnosis.md}, Step 1.} Although the control space is
6-DoF and the state carries RPY, the trained output slot and the success
gate are \textbf{position-only}
($\|$EE$-$target$\|$ $<$ 0.05 m); no RPY term is scored (Amendment 3 \S4).
``6-DoF'' thus implies no rotational success dimension. L2 is \textbf{a second, harder task in the
same primitive family}, neither a push task nor a different skill (a genuine
push, with the cube as a tracked, displaceable object, is L3 scope), and the
task on which the D11 headline is tested for cross-task replication.

\subsection*{5.2 Per-arm scoring (Amendment 1 \S1, disclosed as post-hoc)}

Joint success (both arms within tolerance simultaneously at episode end)
was \textbf{1/100} on the L2 collect; per-arm success was \textbf{28/100 left,
28/100 right}. Each arm reaching its goal at some point
(best-in-episode; 28/100 per arm, both arms in 10/100) is a different event
from both holding tolerance at episode end (1/100), so the
primary L2 metric is \textbf{per-arm} success. Promoting per-arm
scoring to primary is a \textbf{post-hoc} change (Amendment 1 \S1, same
immunisation standard as D11 Amendment 14), motivated by a \emph{structural}
finding about the success definition, not by selection on a protocol
result: no L2 arm had been trained or evaluated at filing time. The joint-success SR is still reported as the originally registered
metric; per-arm is \emph{added, not substituted}.

\begin{itemize}
\item \textbf{A per-arm episode is desirable for arm X} iff arm X's end-effector
  reached its pose goal (best-in-episode $\|$EE$-$target$\|$ $<$ 0.05 m, the
  registered threshold), giving $\approx$28 desirable-arm instances per arm.
\item \textbf{SFT target:} the progress rounds of the desirable arm, with
  canonical action lines emitted for a \textbf{single arm} (only the scored
  arm's target line and STOP), as a single-arm-scored student should
  emit. The non-scored arm's outcome is recorded as
  a flag (\texttt{other\_arm\_reached}), not folded into the target.
\item \textbf{KTO pairing} is per arm instance: desirable examples are the
  scored arm's progress rounds in episodes where it reached its goal,
  undesirable ones its rounds where it failed, mirroring D11's split one arm
  at a time. Left-scored and right-scored episodes both enter the pool.
\end{itemize}

\subsection*{5.3 Buffer re-tag: arm-aligned success-floor (Amendment 1a)}

The 100 L2 recaps were tagged \texttt{is\_success} by joint outcome
(1 True / 99 False). A full re-score found 46/100
episodes with $\ge$1 arm reaching goal, and
\textbf{45 of these carry \texttt{is\_success=False} despite holding a
goal-reaching arm trajectory}. Per-arm, the labels are thus wrong for $\approx$45\% of episodes. The naive fix
(\texttt{is\_success = left OR right}) reintroduces cross-arm contamination:
a right-reached, left-failed episode, whose failure lesson concerns the
left arm, would be served as a success to a left-arm student. The fix has two minimal layers:

\begin{enumerate}
\item \textbf{Data layer.} Each recap gains boolean \texttt{left\_reached} /
   \texttt{right\_reached} fields from replay ground truth;\footnote{Re-score: \repopath{workspace/l2\_audit/per\_arm\_rescore.json}.} joint
   \texttt{is\_success} is kept for reference. Retrieval \emph{similarity is
   unchanged} (image plus full-state anchor, as before).
\item \textbf{Retriever logic.} The success-floor filter reads the label of
   the arm currently scored
   (\texttt{left\_\allowbreak reached}
   when the canonical action line is the left arm's, else
   \texttt{right\_reached}), a lookup of an existing per-round
   variable, not new retrieval logic.
\end{enumerate}

A limitation remains: \emph{lesson text is still episode-level natural
language and may refer to either arm; only the success-floor label is
arm-aligned, not the lesson prose.} Amendment 1a fixed the order of
operations: re-tag, make the 3--5-line retriever change, re-pin the buffer
SHA, and only then run C\_retrieval-L2. Recorded pins: pre-retag
tree hash \texttt{83337879\ldots}, post-retag \texttt{a79581cd\ldots}; 100/100
recaps carry both labels. The teacher's joint-labelled
\emph{growing} buffer vs eval's re-tagged \emph{frozen} buffer asymmetry is
documented in the repository.\footnote{\repopath{docs/paper/l2\_memory\_world\_asymmetry.md}.}

\subsection*{5.4 Scope: L2 tests the answerable question (Amendment 2)}

L2's desirable pool is 511 per-arm rounds from 56 per-arm
episode-instances (L0a: 992 desirable rounds). The two D11 headlines differ
in detectability at this n, so L2 tests \textbf{one} question rather
than a symmetric 2$\times$2 (Amendment 2):

\begin{itemize}
\item \textbf{T4-class (identical-weights retrieval, C\_retrieval $-$ A\_ctrl\_rat).}
  D11's strongest result (+34 pp, z=5.15). At L2 baseline $\approx$15\%,
  power at n=100, $\alpha$=0.010 is \textbf{76\% (+20 pp) /
  99\% (+30 pp) / 100\% (+34 pp)}: detectable, so this is the question L2
  can answer.
\item \textbf{T1-class (bake-in).} L0a effect +1 pp; the L2 critical
  difference at $\alpha$=0.020 and n=100 is \textbf{12--13 pp} (the
  $\approx$50\%-power MDE). A $\approx$1 pp effect on an instrument with a 12 pp
  critical difference and about half the desirable training rounds is a
  guaranteed uninformative null, so T1 is \textbf{not run on L2}: an MDE-driven \emph{design choice} (the
  power analysis applied prospectively to decide what not to run), not an
  omission.
\end{itemize}

\textbf{Stage 1} (this experiment's scope) trains a single adapter,
\textbf{A\_ctrl\_rat}, on all 56 desirable episode-instances. As for training
adequacy (as distinct from evaluation power), the 56 instances expand to \textbf{511 desirable / 1119 undesirable per-arm
rounds}; KTO trains on rounds, so SFT uses 511 behavior-cloning rounds and
KTO uses 511 + 1119. We evaluate two protocols: A\_ctrl\_rat bare, and
C\_retrieval (A\_ctrl\_rat's weights plus the frozen re-tagged buffer,
\texttt{success\_label\_arm='left'}). The primary L2 contrast is
\textbf{C\_retrieval $-$ A\_ctrl\_rat} (identical weights), tested two-sided
with McNemar or z as in D11.

Conditional-expansion criterion (fixed before data, Amendment 2): if
C\_ret $-$ A\_ctrl\_rat is significant and in L0a's direction, the cross-task
headline is secured and A\_action\_only may optionally be added
(Stage 2) for the L2 ``rationale tax''; D\_gist runs in neither stage. If not significant or reversed, that is itself a finding (the effect does not cross tasks), and the response is diagnosis,
not more arms.

The paper's claim is scoped more narrowly, pre-registered as a
\textbf{conditional}: Amendment 2 pre-specified that \emph{a significant,
same-direction contrast would constitute cross-task replication of the
identical-weights retrieval effect} (not of the full 2$\times$2). The
observed contrast (\S5.9.1) did \textbf{not} meet this criterion (direction
agrees, significance does not), routing to Amendment 2's diagnostic branch
(\S5.9.3). The paper thus claims the
mechanism-and-direction reading of \S5.9.3, not effect-size replication.

\subsection*{5.5 Per-arm evaluation: single-arm inference matches single-arm training (Amendment 3)}

Amendment~1~\S3 defined the per-arm \textbf{training} target (scored arm
only); Amendment~3 matches the \textbf{evaluation} to it, under the principle: \textbf{the student is evaluated in the shape
it was trained to emit.}

The evaluation was initially wired to parse, step and gate on \textbf{both}
arms (the level's active-arm resolver returned \texttt{None} for
\texttt{L2\_dual\_push}, selecting the bimanual path), the joint
definition that produced SR = 1/100 and motivated the per-arm re-score.
Asking a single-arm model for both arms measures
\emph{format adaptation, not task competence} (an out-of-distribution
probe). The per-arm eval resolves this:

\begin{enumerate}
\item \textbf{The prompt} asks only for the scored arm's slot, matching the
   training target in structure (position only; no RPY or non-scored-arm
   line).
\item \textbf{Non-scored arm frozen at its initial pose} by the existing V4 \texttt{active\_arm} hold-in-place mask.
\item \textbf{Success gated on the scored arm only} ($\|$EE\_scored $-$ target\_scored$\|$
   $<$ 0.05 m), reusing the L0a single-arm success branch.
\end{enumerate}

\textbf{Freezing the non-scored arm is a structural advantage rather than a
workaround.} A left-scored L2 episode (left arm moving, right arm frozen) is
\emph{isomorphic to the L0a-Left layout}, giving the L2$\leftrightarrow$L0a
comparison the same single-active-arm control structure on a harder task. \textbf{Disclosed difference:} the non-scored arm moved during
training and is frozen at eval. This is defensible for this
position-goal reach \emph{specifically} because the two arms drive to two
\textbf{separate} goals with minimal inter-arm coupling, so freezing one does
not change the physics the scored arm must solve. It would \textbf{not} be
defensible on a contact-coupled cooperative task (L2 is not one; \S5.1).

\textbf{Scope: main eval left-scored only, 2 protocols at n=100 each, shared
seed\_base 4600} (preserving the paired McNemar
design). Left-only is a \emph{scope decision, not a compute shortcut}. Amendment 1a already anchored the retrieval side to the left arm
(\texttt{success\_label\_arm='left'}, retrieval query keyed on the initial
left-EE state), so a right-scored C\_retrieval would need a
\emph{different retrieval anchor}, a new degree of freedom and a new
amendment, which L2 does not open. The left-scored contrast
is a complete, valid test of the identical-weights retrieval question
(Amendment 2). Any right-scored eval is limited to an
\textbf{exploratory bare-arm re-check} (A\_ctrl\_rat only, no anchor change)
run after the main numbers in idle windows, outside the confirmatory
contrast.

\subsection*{5.6 Analysis rules (carried from D11 \S3.6, per-arm generalisation)}

The rules are unchanged in form from D11 (two-timestamp discipline;
pairing-integrity gate, with McNemar primary on pass and two-proportion z on
fail; all tests two-sided). The L2
generalisations were pre-specified in an analysis-adaptation dry run whose
commit timestamp precedes any L2 five-protocol result:\footnote{\repopath{docs/paper/l2\_analysis\_adaptation.md}; its commit timestamp is the pre-specification.}

\begin{itemize}
\item \textbf{Success gate}, per-arm: the
  \texttt{outcome=='success'} extraction reads the scored arm's
  \texttt{dist\_red < thr} (already present in the \texttt{active\_arm=None}
  branch of the evaluation loop); the binary success-count machinery is unchanged.
\item \textbf{Pairing gate.} The seed-based initial-state fingerprint now
  includes both arms (richer \texttt{allclose}, still valid).
  The L2 eval seed base is \textbf{4600}, shared across both protocols,
  so episode \emph{k} starts from the same world configuration in A\_ctrl\_rat
  and C\_retrieval (paired design, Amendment 3 \S3). It differs from
  D11's eval seed base 4500 (the L2 teacher collect was unseeded).
\item \textbf{R1-bias probe (L2):} a pre-specified redefinition (\S5.7).
\end{itemize}

\subsection*{5.7 R1-bias probe, L2 redefinition (locked before data)}

This redefinition was locked before any L2 data, in the same
pre-specification file as \S5.6. The L0a R1 probe measured the single active arm's round-1 lateral (X)
bias. The pre-registered L2 equivalent is the \textbf{per-arm round-1
signed displacement toward each arm's own target, reported as a
two-component vector ($\Delta$X\_L, $\Delta$X\_R), plus a pooled magnitude
$\|\Delta$R1$\|$} (mean over both arms of $|$round-1 target $-$ init
EE$|$ along the task-relevant axis). The marginal-vs-conditional test R1
served in L0a carries over unchanged in form: does the distilled
arm's per-arm R1 collapse to a static prior while retrieval's varies per
episode? References are the \textbf{L2 memory teacher's own} R1
distribution (task-matched, measured from the L2 teacher buffer),
\textbf{not} the transplanted L0a $-$23.5/$-$15.8 grid-unit
fingerprints. It was locked before results so it cannot be a metric
picked after the fact.

\subsection*{5.8 Pipeline integrity (verify-layer interceptions, carried from D11)}

The D11 Step-2 row-count sentinel and SHA gates carry over. The L2 run
exercised the verify layer twice more (interception rate 3/3 across the
program). (i) The per-arm SFT pool lies almost entirely in the
\texttt{failure/} replay dir (joint success was 1/100), so the pool is
selected by per-arm \emph{label} (\texttt{--sft\_desirable\_only},
kto\_label=='desirable') reading both dirs, not by directory. The
sentinel caught a would-be training-semantics error (1119 failed-arm
rounds nearly behavior-cloned as desirable), not a miscount (Amendment 2
addendum 2026-07-16). (ii) A version drift in the GGUF conversion toolchain
transposed the LoRA storage layout; it was caught at the GGUF-load verify
gate and fixed by a version rollback plus a structural architecture
byte-patch, not by hand-transposing tensors, which risks a load that
succeeds but silently mis-maps (Amendment 2 addendum 2026-07-21).

A fourth interception (Amendment 3 \S7) closes a recurring error family.
The L2 eval initially parsed both arms for a single-arm-trained model:
the fourth train/eval output-format mismatch caught in the
program, but caught \emph{reactively} (a crash at episode 1). The
permanent fix is a \textbf{format-contract assertion} in the driver, run
\emph{before} any eval collect: verbatim training-target rows from the
SFT JSONL are fed through the eval parser configured exactly as eval will
configure it (control mode, variant and
\texttt{scored\_arm}), and every row must parse. A single-arm target
fed to a both-arms parser fails this in a dry run, before the simulator
starts. The train/eval-contract
interception rate is thus 4/4, and prospective rather than reactive from
here on.

\subsection*{5.9 Results}

The analysis rules were locked before any $p$-value (pre-analysis lock).\footnote{Confirmatory analysis: \repopath{scripts/analysis/l2\_confirmatory.py}; run IDs: A\_ctrl\_rat \texttt{8384a740}, C\_retrieval \texttt{2154e57e}, L2 teacher \texttt{a6e6c917}.}

\subsubsection*{5.9.1 Confirmatory contrast: the identical-weights retrieval effect}

Per-arm eval (Amendment 3: left-scored, n=100 paired, seed\_base 4600):

\begin{table}[h]
\centering
\begin{tabular}{l l l}
\toprule
protocol & successes / n & SR \\
\midrule
A\_ctrl\_rat (bare) & 14 / 100 & 14\% \\
C\_retrieval (same weights + frozen re-tagged buffer) & 20 / 100 & 20\% \\
\bottomrule
\end{tabular}
\end{table}

The identical-weights contrast is \textbf{C\_retrieval $-$ A\_ctrl\_rat = +6.0 pp}
(Newcombe 95\% CI \textbf{[$-$4.5, +16.4] pp}). Primary test: two-proportion
z = 1.129, \textbf{p = 0.259, n.s.} at the pre-registered $\alpha$=0.010
(T4-class). The paired McNemar was the intended primary; the
pairing-integrity gate fell back to z per Amendment 14 \S14.2 (\S5.9.2), and
McNemar's discordant cells (9 vs 3; exact McNemar p=0.146) give the same
n.s. conclusion.
Auto-reset sensitivity (Amendment 15 pin 8 rule, \S3.5; L2 inherits the
same 24\,s time-out): 10 \texttt{A\_ctrl\_rat} and 9 \texttt{C\_retrieval}
episodes cross the boundary and none succeeds after it, so the
sensitivity counts equal the as-run counts: 20 vs 14, +6.0 pp (z=1.129,
p=0.259), same verdict.\footnote{\repopath{docs/paper/data/l2\_autoreset\_sensitivity.json}.}

The pre-analysis lock states the verdict as follows: \emph{the point
estimate is below the pre-registered MDE (+20 pp for 76\% power, Amendment
2); the direction agrees with L0a; the confidence interval simultaneously
admits zero and a moderate effect.} By Amendment 2's pinned rule, a
sub-MDE estimate triggers the \textbf{diagnosis branch} (\S5.9.3), not
more arms. The +6 pp result was predicted in advance to be n.s.,
and it is.
We do \textbf{not} call +6 pp a positive replication of the effect
\emph{size}; it is n.s. and sub-MDE. What points the same way is examined
mechanistically below (exploratory).

\subsubsection*{5.9.2 Pairing integrity (why z, not McNemar)}

The environment seed matched across protocols on 100/100 episodes, and the frozen right arm (Amendment 3 hold-in-place) matched at
integer and RPY resolution on 94/100. The float fingerprint gate
($\varepsilon$=1e-4) failed on 17 episodes at $\approx$1e-4 drift:
\texttt{trajectory[0]} is already one servo step in and the scored (left)
arm's differing first action perturbs the shared simulator's float state.
Pairing is \emph{physically real but not machine-verifiable from the
persisted replay}, so the mechanical rule (Amendment 14 \S14.2) selected the
two-proportion z-test. D11 disclosed the same fallback, and the result is n.s.
either way. No discretion was exercised.

\subsubsection*{5.9.3 Diagnostic appendix: why the effect is small, not absent}

Each of three pre-registered diagnostics had its prediction committed
first; all three findings are \textbf{exploratory}, and none is
a sole-cause claim.

\textbf{(a) Retrieval quality: a cross-task correlate (n=2 tasks), not a
within-task predictor.} With an identical retriever, L2 retrieves
systematically less similar neighbours than L0a (top-1 median 0.900 vs 0.870, MW p=4.7e-17; top-3 0.883 vs 0.827, p=1.3e-26), consistent
with a 5.5$\times$ thinner buffer (547 vs 100 recaps, $\approx$28 usable
left-aligned). Stated at the level it holds, \emph{the L2 attenuation is
consistent with reduced buffer coverage}: a correlation across two tasks, not a causal ``$\propto$'', and explicitly \textbf{not} a claim
that match quality predicts which L2 episodes succeed (see (c), where it
does not).

\textbf{(b) Retrieval restores the teacher's round-1 bias.} Round-1 left-arm
lateral bias $\Delta$X\_L, against the L2 teacher's \emph{own} R1
rather than L0a's transplant: teacher $-$17.1, A\_ctrl\_rat
(distilled) \textbf{$-$4.8}, C\_retrieval \textbf{$-$17.2}. The
pre-registered $\sigma$ test did not fire ($\sigma$ 15.5 vs 17.3, F=1.25).
The informative signal is the \textbf{mean}: distillation collapses to a
magnitude-deficient prior ($\approx$28\% of the teacher's displacement), and
the retrieval preamble on identical weights restores the teacher's magnitude
almost exactly ($-$4.8 to $-$17.2, teacher $-$17.1). This
is the L2 analogue of the L0a static-prior collapse (D11 \S4.3), on the mean
axis (mean finding post-hoc and exploratory; the $\sigma$ test was the
registered one).

\emph{Exploratory cross-task note: even the marginal transfer appears
dose-dependent.} On L0a the distilled arm's R1 was close to the teacher's
($-$16.7 vs $-$15.8); on L2 the distilled prior reaches only $-$4.8 against
$-$17.1 ($\approx$72\% short). Training dose is one candidate (L2's per-arm
distillation saw 511 desirable rounds, L0a's 992). It suggests that
\textbf{marginal transfer is cheap but not free}: below some data floor even
the static prior is under-learned. But task difficulty and buffer
composition are confounded with dose at n=2, so no single cause is
isolable from two tasks and a single before/after point.

\textbf{(c) Rescue is gated by start distance, not memory match.} The +6 pp is 9 rescues (C\checkmark A$\times$) minus 3
regressions (A\checkmark C$\times$). Rescues share a sharp init-pose
signature: mean init \texttt{dist\_red} 13.9 cm for rescues vs 21.6 cm for
both-fail (MW z=$-$3.88, p=1e-4), so retrieval rescues
almost only close-start episodes. Rescue similarity is
\emph{not} higher (0.847 vs 0.868, p=0.03), so start distance, not match
quality, is decisive within L2. This was not the expected outcome: the expectation
recorded with the pre-registered prediction was a diffuse,
distance-independent rescue set.

\textbf{Synthesis.} The identical-weights retrieval effect is consistent on
L2 in \textbf{mechanism} (the R1 correction in (b); post hoc, since the
registered $\sigma$ test was null) and in \textbf{direction} (+6 pp), but
its conversion to end-task success is \textbf{competence-gated}. The
retrieved correction is injected \textbf{once per episode} (anchored on
\texttt{init\_L\_EE}, round 0 only). It
biases the initial direction, nothing re-aims it mid-trajectory, and so it
converts only where the start is already within reach of the 0.05 m gate
(c). Far-start episodes get the correction yet miss. The effect is
thus \textbf{small (+6 pp, n.s.), not absent.} Two levels are kept separate,
answering the objection that similarity explains attenuation but not
rescue: \emph{between tasks} (a, n=2), a thinner
buffer co-occurs with a smaller effect (its \textbf{size}); \emph{within L2}
(c), conversion is gated by start distance (its \textbf{incidence}).

\subsubsection*{5.9.4 Headroom-normalized cross-task framing (post-hoc, exploratory)}

Raw pp understate the effect across tasks of different headroom. On a
gap-to-teacher basis, L0a retrieval closed $\approx$99\% of the parity gap
(34.0 of 34.3 pp; A\_ctrl\_rat 15\% against a teacher at 49.3\%), while
\textbf{L2 retrieval closes 42.9\%} of the per-arm gap (A\_ctrl\_rat 14\%
against the L2 per-arm teacher's \textbf{28.0\%}, pinned from the per-arm re-score; headroom 14 pp, +6 pp recovered,
6/14). This framing is \textbf{post-hoc and exploratory}
(introduced after the raw SR was known). Pinning the 28.0\% denominator
removes the ``$\approx$'', not the exploratory label; the raw +6 pp n.s. is
always reported alongside.

\section*{Appendix: L2 amendment chain (extends D11 Appendix A1)}

\begin{table}[h]
\centering
\small
\begin{tabular}{@{}l p{1.7cm} p{6.4cm} p{3.1cm}@{}}
\toprule
\# & Date & Change & Anchor \\
\midrule
L2-1 & 07-15 & Per-arm scoring (post-hoc, disclosed); SFT single-arm target; task rename & l2\_amendment\_1.md \\
L2-1a & 07-15 & Buffer re-tag (dual label) + arm-aligned success-floor; pins 83337879$\rightarrow$a79581cd & l2\_amendment\_1.md \S1a \\
L2-2 & 07-15 & Scope to answerable question (T4-class only); Stage-1 A\_ctrl\_rat; T1 not run (MDE design choice) & l2\_amendment\_2.md \\
L2-2 add & 07-16 & Per-arm SFT pool selected by label not directory (verify-layer 2/2) & l2\_amendment\_2.md \\
L2-2 add & 07-21 & GGUF conversion version-drift caught at load gate; rollback + arch byte-patch (verify-layer 3/3) & l2\_amendment\_2.md \\
L2-3 & 07-22 decided / 07-28 re-filed & Per-arm EVAL: single-arm inference matches single-arm training; non-scored arm frozen; left-scored $\times$2 protocols n=100, seed 4600; position-only success; format-contract gate (verify-layer 4/4) & l2\_amendment\_3.md (\commit{96e1388}; spec-implementing code \commit{0398568}, 07-22, before the eval) \\
\bottomrule
\end{tabular}
\end{table}

\emph{L2-3 was decided before any L2 eval number existed: the Step-8
eval crashed at episode~1 on a missing prompt template, so the
per-arm eval specification was fixed before the eval was restarted and its
numbers observed. This preserves the pre-registration
ordering (specification precedes results) for the per-arm eval, as for the
rest of the plan.}

\section*{6. Discussion}

\subsection*{6.1 What the results say to a system builder: externalise the memory}

The two headline results are one finding seen from two sides. Baking
the memory-augmented teacher's behaviour into student weights bought
+1 pp over an action-only control (T1, null; sensitivity $-$2 pp); attaching an
inference-time retrieval context to a \emph{fixed} set of student weights
bought +34 pp (identical-weights contrast, z=5.15; the registered
protocol contrast agrees, +23 pp, z=3.36; sensitivity +36 / +29 pp) and
reached the pooled memory-teacher reference (49\% vs 49.3\%; sensitivity 48\% vs 47.5\%). Competence tracks the teacher at each memory rung (\S4.2); what
distillation fails to move into weights is specifically the memory
advantage.

We resist reading this as ``distillation fails''. The exploratory
mechanism probe (\S4.3) suggests \emph{why} the split is principled, and
turns it into a design criterion. Memory's operational value here is a
\emph{conditional} correction: a different adjustment for each episode's
situation (the C\_retrieval R1 $\Delta$X signature: mean pulled toward zero ($-$7.1), $\sigma$=14.4,
per-episode variable). Baking a conditional function into a fixed
weight delta forces a single static approximation of it; the four
distilled arms all collapsed onto one $\approx-$16.7 (grid-unit) prior regardless of the
episode in front of them. A context channel, by contrast, is a natural
conditioning pathway: it can present a different retrieved context (a
past episode's lesson and image) per situation at inference with no
parameter change.

The answer to ``where should this knowledge live'' therefore depends on
the knowledge's own structure. \textbf{Marginal knowledge (the
average correction that holds across situations) transfers into
weights; conditional knowledge (the correction that depends on the
situation) belongs in context, because weights approximate it with a
constant and context conditions on it for free.} This is the criterion
D11 was ultimately built to test, and it is the transferable claim we
carry out of a single-recipe study across two tasks (the L0a reach and
the harder L2 position-goal reach in a 6-DoF control space, \S5.9.3).
The claim is not that retrieval beats baking-in as a general law; it is a
decomposition of \emph{which} component of an experience each substrate
can hold.

What the recipe does not buy at our scale is speed. Teacher and student
share the 31B base and both generate a rationale before the action, so
per-step cost is of the same order for all three protocols, and decode
time dominates each step (\S4.5). A latency advantage requires a smaller
student or a learned action decoder that emits actions without
generating text. The decomposition says what such a decoder should hold:
the marginal prior can move into it, while the memory stays in context.

\subsection*{6.2 Two poles, a bridge, and a loop}

The literature splits into two substrate poles; our contribution is the
bridge between them, and the open frontier is the loop that closes over
both.

\textbf{Pole 1: external memory (we supply supporting evidence).} The
retrieval-augmented pole, in which competence is distilled and experience
is kept external and looked up, is the world-view our results support: on
\emph{fixed} student weights, attaching retrieval added +34 pp (sensitivity +36)
(identical-weights contrast, z=5.15; the registered protocol contrast
agrees at +23 pp). Embodied retrieval-augmented policies already exist:
behaviour/skill retrieval and episodic-memory agents (Voyager~\citep{voyager2023},
ExpeL~\citep{expel2024}, AWM~\citep{awm2024}; ASPIRE~\citep{lu2026aspire})
show external-memory gains, so we do not claim primacy for the idea.
What we add is a \emph{pre-registered} embodied comparison that isolates
substrate: the same retrieved memory routed into weights vs into context,
with the exploratory mechanism probe suggesting \emph{why} the routing matters (marginal
vs conditional). ASPIRE's code-as-reusable-skill and our
lesson-and-image retrieval are two instantiations of external memory; our data
support the pole, not any single implementation.

\textbf{Pole 2: parametric memory (we bound our null and leave the
question open).} Our T1 null is a statement about a \emph{behavioural distillation
recipe} (single-round SFT + composable-KTO, LoRA), not about parametric
memory in general. Titans-style architectural writing~\citep{titans2025}
(surprise-gated updates that write only the unexpected, with decay)
and parametric self-evolution (TMEM~\citep{tmem2026}) are
mechanisms our recipe lacks: they consolidate \emph{conditionally} (what to
write depends on the situation's surprise), which is exactly the component
\S4.3 shows our recipe drops. MemHarness~\citep{memharness2026} makes the
same point from the other side, treating memory as \emph{reconstructed}
rather than replayed. Whether such a mechanism can bake the conditional structure that
flat distillation could not is untested here and is the most direct
architectural follow-up our null points to. The case for it is
strengthened because the data route is not closed either: our no-training
diagnostic excludes a conditional trace only at round 1, while later rounds
show a positive but confounded signal (\S4.4b). A conditional component
may be present but under-consolidated by a flat recipe, which is the
regime surprise-gated (Titans) and self-evolving (TMEM) writers target.

\textbf{Provenance caveat (an open question).} C\_retrieval is fed the
\emph{teacher's} frozen experience buffer, not the student's own. For the
question D11 asks (the \emph{same} memory, routed to weights vs
context), buffer source is a controlled constant (B\_main was distilled
from trajectories the teacher produced while retrieving from the frozen
buffer, and C\_retrieval retrieves from that same buffer), so the buffer is
held fixed across the substrate contrast. What reaches the student differs
in form: B\_main's target carries the lesson gist as text, whereas
C\_retrieval's context carries each retrieved lesson with its episode
image. What we do \emph{not} test is whether a student
retrieving its \emph{own} self-generated buffer would win by the same margin;
self-produced recap quality is unknown and connects to the epiphenomenal
self-gist finding (\S4.3). We flag this as future work rather than a resolved
claim: externalisation as the mechanism and teacher memory quality can be
separated only by a student-own-buffer arm, which we have not run.

\textbf{The bridge: which component of an experience each substrate
carries.} The two poles are usually argued as rival systems; our data turn
the choice into a decomposition. Marginal structure (the average
correction) consolidates into weights; conditional structure (the
situation-specific correction) is what retrieval supplies and distillation
drops (\S4.3). This is substrate-agnostic: it predicts \emph{what} a
parametric mechanism would have to capture (the conditional component) to
match retrieval, which is what the surprise-gated and self-evolving
writers above target. The criterion connects the two poles rather than
choosing one. More speculatively, if the criterion holds, it predicts the
apparently opposite finding of \citet{routing2026}, where lessons did better
in the weights: lessons that state situation-independent rules are marginal
knowledge and should favour the weights, whereas ours are per-situation
corrections, retrieved together with their episode images, and favoured
context.

\textbf{The loop: closing experience back into the policy.} The frontier
is agents that keep improving by folding their own rollouts back in, as in
closed-loop iterative fine-tuning (CLIFT~\citep{clift2026}) and
self-evolving parametric memory (TMEM). Our result bears on that
loop: it suggests the loop's \emph{write} step (consolidation) loses the
conditional component unless built to preserve it, so a naive
rollout-then-distil loop would be expected to plateau at the marginal. A
follow-up paper (J-Space / J-lens, in preparation) develops the instrument
for the mechanistic half: our exploratory R1 $\mu$-branch result (\S4.3) suggests
behaviourally that the correction's carrier is the action distribution,
not the memory content (\texttt{B\_main} still emits PAST\_LESSONS text
yet it bought +1 pp, suggesting it is epiphenomenal). Ablating
\texttt{B\_main}'s self-produced gist (prediction: action unchanged), and
ablating the text versus the image component of \texttt{C\_retrieval}'s
retrieved context (which the present design cannot separate), would give
a mechanistic account of the 23 pp gap and identify which component carries
the conditional correction. This is the target of the follow-up paper and
is not a result claimed here.

\subsection*{6.3 Why this does not extend to contact-rich manipulation (a bounded feasibility probe)}

Before extending the memory result to contact-rich tasks we ran a
bounded feasibility gate, porting the identical prompted-VLM-teacher
interface to LIBERO~\citep{libero2023} and asking whether it clears the task before
committing a full study. It did not, and the negative result is precise. Across grasping (three objects) and non-grasping
(push, knob-twist) tasks the teacher's \emph{spatial reasoning stayed intact}
(round-0 targets matched true object positions to within millimetres
and phase identification was correct throughout), yet success was 0 in
every task. A staged intervention (adding oracle-sourced subgoal
vectors, then a hover-descend-grasp primitive, then raising camera
resolution from 128 to 256 px; details and per-stage success in the
supplementary feasibility branch) cleared every candidate cause above
the point of contact while success remained pinned at zero, which points
to two separable boundaries (exploratory). The first is a
\textbf{contact-precision ceiling}: on free-object contact (grasp and push
alike) the open-loop position servo drives the end-effector into the
object and the rigid-body contact impulse ejects it, so the same wall
that blocks grasping also blocks pushing. The second is a
\textbf{primitive-expressivity gap}: the ``turn on the stove'' success
predicate is satisfied by rotating a hinged knob past $\approx$0.5 rad, a
degree of freedom our macro vocabulary (reach / grasp / release) does
not express, so a position servo to the knob applies zero moment about
the hinge and cannot actuate it even with perfect targeting. Both
boundaries are properties of \emph{this interface paradigm (integer-coordinate
subgoals over an OSC position servo with a fixed macro-primitive set),
not of the prompted-VLM teacher}, whose reasoning layer
was sound in every failure; the walls sit at the control-stack and
primitive-vocabulary layers, the layers this single-manipulator study
does not claim to solve. This bounds the contribution and indicates the
next step: \textbf{closed-loop contact control and an expanded primitive
vocabulary are the two extensions} a richer task suite would require. An
expert-iteration loop (the follow-up paper) would need the same enrichment
to generate contact-rich experience worth externalising.

\subsection*{6.4 The open problem: consolidating a conditional function, not a marginal prior}

The broadest reading connects back to the study's motivating question:
whether the knowledge a system reads can be folded, through embodiment,
into its parameters. This is the systems-consolidation question in
machine form, the slow transfer of a fast episodic store into
structured, automatic competence that complementary-learning-systems
theory describes~\citep{mcclelland1995complementary, kumaran2016learning}.
Human skill automation is the existence proof that it can: a fluent
pianist's playing is automatic yet still \emph{conditional}, responding
note-by-note to what is in front of them. Automation preserves the
conditional structure; it does not collapse the skill to its average.
Our distillation product kept only the average.

That reframes the open problem. T4 appears to settle ``baked-in vs
retrieved'', but it settles only the question of substrate under one
consolidation recipe. The problem underneath is \textbf{how to consolidate
a conditional function rather than a marginal prior}. That single question explains all three of our
results at once: T1 failed because flat distillation consolidates the
marginal; T4 succeeded because context still carries the conditional; and the
paths beyond (surprise-gated writing, per-round distillation targets
that supervise the situation-specific correction rather than its
average, and workspace-readout probes that check whether a consolidated
correction is actually used) are all attempts on that same problem.
The null on substrate thus reformulates ``where should memory live'' as
``what would it take to consolidate memory's conditional structure at
all'', a question embodied continual learning has yet to answer.

\subsection*{6.5 The cross-task result (Experiment 2, L2)}

The second experiment (\S5) tests whether the identical-weights retrieval
effect crosses to a harder task. It connects to the substrate account in four
ways, each tied to the pre-registered analysis:

\begin{itemize}
\item \textbf{The marginal/conditional split on a second task (direction-consistent, post hoc).}
  L2 is consistent with the \emph{mechanism}: distillation transfers a
  magnitude-deficient marginal prior (R1 $-$4.8), and retrieval on identical
  weights restores the teacher's conditional correction ($-$17.2 $\approx$ $-$17.1;
  post-hoc, as the registered $\sigma$ test was null, \S5.9.3b). The end-task
  effect stays below the MDE (+6 pp, n.s.). The substrate account points the same
  way in mechanism and direction, which is not a replication; the effect
  \emph{size} does not reach significance on this harder task with its
  thinner buffer.
\item \textbf{Marginal transfer is dose-dependent (exploratory).}
  L2's $\approx$72\%-short marginal (\S5.9.3b) adds a dimension to the marginal/
  conditional frame: even the static prior has a data floor below which it
  is under-learned. It is cheap but not free.
\item \textbf{The memory success-floor / minimum-viable-competence law.}
  L2 at 1\% joint SR sits near the floor: memory
  conditions the correction but does not manufacture the multi-step
  competence to execute it from far configurations (\S5.9.3c). \emph{You cannot
  externalise a competence the teacher does not have.} Here the statement
  sharpens to: \emph{retrieval sets the initial direction; whether that pays off depends on
  how far the corrected trajectory still has to travel.} We state this as
  an exploratory hypothesis for a follow-up study, unifying observations
  across the teacher, the buffer and the student; it is not a result of this paper.
\item \textbf{A link between the two experiments (exploratory).} This matches
  D11's recovery-timing null: both datasets suggest retrieval's value is
  \textbf{front-loaded}: it sets the initial direction (L2's start-distance
  gate is its range limit; D11's monotonic-from-R1 convergence is its
  timing signature).
\end{itemize}

Note the contact-rich boundary (\S6.3) and this cross-task result are
independent: \S6.3 bounds where the \emph{interface} can go, \S6.5 reports
how the \emph{memory} finding travels within the reach family.

\appendix
\section*{Appendix A. Pre-registration amendment chain}

The full amendment chain is tabulated with the Methods. The D11
amendments (1--14) are in Appendix~A1 at the end of the Experiment-1
Methods (\S3). Amendment 15 (2026-10-05) registers inference-only
reinforcement arms that post-date the D11 results. Amendment 16
(2026-10-05, \commit{730b9fc}) is a documentation-only erratum that
relabels R1 $\Delta$X from cm to scalar-guard grid units, with no
analysis change. The L2 amendments (1/1a/2/3) are in the table at the
end of the Experiment-2 Methods (\S5). Every amendment is signed, dated, and anchored
to a frozen commit in the pre-registration document. Derived statistics
not printed by the registered analysis scripts (Newcombe intervals, the
identical-weights $z$, the T1 critical difference, the pooled teacher
rate) are recomputed from persisted counts by
\repopath{docs/paper/birth\_certificates/derived\_stats.py}.

\section*{Teacher configuration (D10/L0a and L2 collects)}

\subsection*{Shared configuration}

The teacher launch script \repopath{server/llama\_server\_teacher.sh} carried
\texttt{--reasoning-budget 512} unchanged from commit \commit{a2aebb9}
(2026-06-08) through the entire D10 and D10-ext campaign (late June) and
into the first L2 collect (2026-07-14). Both tasks used the same binary,
the same model (gemma-4-31B-it Q4\_K\_M), the same mmproj, ctx-size 16384
and \texttt{--parallel 1}.

\subsection*{Recap outcomes under the shared configuration}

\begin{itemize}
\item \textbf{D10 (L0a-Left, single-arm):} inline Stage-4 recap generation
  succeeded almost completely. The D10 collect logs contain no
  ``Empty content'' recap failures; the isolated empty-recap skips that
  do occur coincide with teacher-server connection outages
  (recaps\_d10 = 547 records total across 7 runs).
\item \textbf{L2 (dual-arm position reach) first collect:} inline recap failed in 8/8 episodes,
  each with ``Empty content in VLM response''.
\end{itemize}

The same teacher configuration therefore gave opposite recap outcomes on
the two tasks. The cause was confirmed deterministically on 2026-07-14:
gemma-4's reasoning mode, enabled by \texttt{--reasoning-budget 512}, can
consume the entire 400-token output budget on \texttt{reasoning\_content}
and return empty \texttt{content} (finish\_reason=length). The reflective
recap prompt triggers this, and the L2 task lengthens that prompt (dual-arm
state, more rounds, a longer round history) enough to push the model into
unbounded reasoning, which the shorter L0a recap prompt did not. The L0a
recap stayed within budget because of the prompt's brevity, not by design.

\subsection*{Change for the L2 collects}

The teacher was relaunched with \texttt{--reasoning off --reasoning-budget 0}
(commit \commit{cabab3c}). The same L2 recap call that had returned empty
then returned a 469-character lesson in 7 s. Reasoning-off is also safe
for L0a: D10 succeeded with budget 512 because it never needed the
reasoning tokens, so removing them changes nothing there.

\subsection*{Implications for comparing L0a and L2}

\begin{itemize}
\item D10 (L0a) recaps were generated with \textbf{reasoning on (budget 512)};
  L2 recaps are generated with \textbf{reasoning off (budget 0)}.
\item This teacher-configuration difference between the two tasks'
  \emph{recap buffers} was introduced by the change above. It does
  \textbf{not} affect the students, each of which is trained from its own
  buffer's text, but it is a provenance difference that applies if the
  L0a and L2 memory buffers are compared directly.
\item Stage-1 reasoning quality may differ subtly between reasoning-off and
  reasoning-on; D10's Stage-1 ran with reasoning on. A clean L0a-vs-L2
  comparison would require either re-running L0a's recap buffer with
  reasoning off for parity, or reporting the difference explicitly.
\end{itemize}

\begin{table}[h]
\centering
\small
\begin{tabular}{@{}lll@{}}
\toprule
Collect & Teacher & Reasoning \\
\midrule
D10 / L0a & gemma-4-31B Q4\_K\_M, ctx 16384 & budget 512 (on) \\
L2 (from 2026-07-14) & same & off (budget 0) \\
\bottomrule
\end{tabular}
\end{table}

\section*{Code and data availability}
\ifanonymised
Code and the pre-registration chain will be released with the camera-ready.
\else
Code, the pre-registration documents and their amendment chain, the
reproduction guide (\url{docs/paper/REPRODUCE.md}) and the
birth-certificate generators are in the public repository
\url{https://github.com/TPEmist/AionGenos}. The two \texttt{A\_ctrl\_rat}
adapters and the frozen retrieval buffer with its 547 \texttt{init\_pre}
images are assets of its GitHub Release \texttt{paper-v1-artifacts}. Files
cited under \texttt{logs/} and \texttt{workspace/}, and the per-run audit
outputs cited by file name (\texttt{exploratory.json},
\texttt{per\_arm\_rescore.json}, \texttt{training\_meta.json}), are local
audit outputs derived from the raw replays and are \textbf{not released}.
\fi

\bibliographystyle{plainnat}
\bibliography{refs}

\end{document}